\documentclass[10pt,journal,compsoc]{IEEEtran}
\usepackage{balance}
\usepackage{ulem}
\usepackage{booktabs}
\usepackage{makecell}
\usepackage{color}
\usepackage{multirow}
\usepackage[citecolor=blue, colorlinks]{hyperref}
\ifCLASSOPTIONcompsoc
\usepackage[nocompress]{cite}
\else
\usepackage{cite}
\fi

\ifCLASSINFOpdf
\usepackage[pdftex]{graphicx}
\usepackage{epstopdf}
\else
\fi

\usepackage{amsmath,amssymb}

\ifCLASSOPTIONcompsoc
\usepackage[caption=false,font=footnotesize,labelfont=sf,textfont=sf]{subfig}
\else
\usepackage[caption=false,font=footnotesize]{subfig}
\fi

\usepackage{dblfloatfix}

\begin{document}

\title{Content-aware Warping for View Synthesis}

\author{Mantang Guo,
        Junhui Hou,~\IEEEmembership{Senior Member,~IEEE},
        Jing Jin,~
        Hui Liu,~
        Huanqiang Zeng,~\IEEEmembership{Senior Member,~IEEE},
        and Jiwen Lu,~\IEEEmembership{Senior Member,~IEEE}
\IEEEcompsocitemizethanks{\IEEEcompsocthanksitem  This work was supported in part by the Hong Kong Research Grants Council under Grant 11218121 and Grant 21211518, in part by the Hong Kong Innovation and Technology Fund under Grant MHP/117/21, in part by the Basic Research General Program of Shenzhen Municipalityunder under Grant JCYJ20190808183003968, and in part by Hong Kong University Grants Committee under Grant
UGC/FDS11/E02/22. (Corresponding author: Junhui Hou)).
\IEEEcompsocthanksitem M. Guo, J. Hou, and J. Jin are with the Department of Computer Science, City University of Hong Kong, Hong Kong, and also with  with the City
University of Hong Kong Shenzhen Research Institute, Shenzhen 518057,
China (e-mails:  mantanguo2-c@my.cityu.edu.hk;  jingjin25-c@my.cityu.edu.hk;jh.hou@cityu.edu.hk).
 \IEEEcompsocthanksitem H. Liu is with the School of Computing \& Information Sciences, Caritas Institute of Higher Education, Hong Kong. E-mail:hliu99-c@my.cityu.edu.hk. 
 
  \IEEEcompsocthanksitem H. Zeng is with the 
  School of Information Science and Engineering, Huaqiao University, Xiamen, China. E-mail: zeng0043@hqu.edu.cn
  \IEEEcompsocthanksitem J. Lu is with the Department of Automation, Tsinghua University, Beijing, China. E-mail: lujiwen@tsinghua.edu.cn 
  
      }
      }


\IEEEtitleabstractindextext{%
\begin{abstract}

Existing image-based rendering methods usually adopt depth-based image warping operation to synthesize novel views.
In this paper, we reason the essential limitations of the traditional warping operation to be  
the limited neighborhood and only distance-based interpolation weights.
To this end, we propose \textit{content-aware warping}, which adaptively learns 
the interpolation weights for pixels of a relatively large neighborhood from 
their contextual information via a lightweight neural network.
Based on this learnable warping module, we propose a new end-to-end learning-based framework for novel view synthesis from  a set of input source views, in which two additional modules, namely confidence-based blending and feature-assistant spatial refinement, are naturally proposed to handle the occlusion issue and capture the spatial correlation among pixels of the synthesized view, respectively.
Besides, we also propose a weight-smoothness loss term to regularize the network. 
Experimental results on light field datasets with wide baselines and multi-view datasets show that the proposed method significantly outperforms state-of-the-art methods both quantitatively and visually. The source code will be publicly available at  \url{https://github.com/MantangGuo/CW4VS}.
\end{abstract}

}

\maketitle

\IEEEdisplaynontitleabstractindextext
\IEEEpeerreviewmaketitle

\IEEEraisesectionheading{\section{Introduction}\label{sec:introduction}}
\IEEEPARstart{N}{ovel} view synthesis aims to generate views that
mimic what a virtual camera would see in between two or more reference views \cite{szeliski2010computer}, which can benefit various downstream applications, such as 3D reconstruction \cite{furukawa2009accurate,schonberger2016structure,yao2018mvsnet,guo2020accurate} and virtual reality \cite{lfapp2015vr,lfapp2017vryu,wei2019vr}. 
Over the past decades, a considerable number of view synthesis methods \cite{debevec1996modeling,levoy1996light,zhou2018stereo,mildenhall2019local,wang2021ibrnet} have been proposed (see Section \ref{sec:related_work} for the comprehensive review).
Particularly, image-based rendering (IBR) methods \cite{debevec1996modeling,levoy1996light,chaurasia2013depth,hedman2016scalable,riegler2020free,shi2021self} perform view synthesis by the depth-based warping operation. 
Generally, they first warp input source views to the novel view based on the estimated depth map, and then blend the warped images to produce the novel view.
These methods mainly focus on improving the depth estimation, the blending strategy, 
or post-processing refinement.

Different from existing works, in this paper, we tackle the problem of novel view synthesis based on an insight that the commonly adopted warping operation confronts with natural limitations.
Specifically, the traditional warping operation synthesizes pixels of a novel view by performing interpolation using a limited neighborhood from source views, and determines the interpolation weights using only  distance-based functions.
The reconstruction quality is thus limited  because the content information around the interpolated pixels, such as texture edges, occlusion boundaries, and non-Lambertian objects, are not considered (see Section \ref{sec:problem_analysis} for the detailed analysis).

To this end, we propose content-aware warping to replace the traditional warping operation, in which a lightweight neural network is utilized to learn content-adaptive interpolation weights for pixels of a relatively large neighbourhood from their contextual information.
Based on this learnable warping module, we construct a new end-to-end learning-based framework for novel view synthesis from a set of input source views.
To be specific, we first warp the source views separately to the novel view pixel-by-pixel via the proposed content-aware warping, and then adaptively leverage the warped views via  confidence-based blending to handle the occlusion problem, leading to an intermediate result of the novel view.
As the pixels of the intermediate result are independently synthesized, we subsequently recover the spatial correlation among them by referring to that of source views using the feature-assistant spatial refinement module.
We also regularize the network through a weight-smoothness loss.

In summary, the main contributions of this paper are as follows:
\begin{itemize}
    \item we analyze the classic 2D image warping operation when used for novel view synthesis and figure out its limitations in terms of the limited neighborhood and only distance-based interpolation weights;
    \item we 
    propose learnable content-aware warping, which is capable of overcoming the limitations of the traditional warping operation; and 
    \item we propose a new end-to-end learning-based framework for novel view synthesis.
\end{itemize}

A preliminary version of this work was published in ICCV’21\cite{guo2021learning}. Compared with the conference version, the additional technical contributions of this paper are four-fold: 
\begin{itemize}
    \item we embed the global content information for learning the content-adaptive weight;
    \item we propose a feature-assistant spatial refinement module; 
    \item we propose a novel weight-smoothness loss term; 
    and
    \item we extend the framework 
    to synthesize novel views on multi-view datasets and experimentally demonstrate its significant advantages over state-of-the-art approaches. 
\end{itemize}

Extensive experiments on both  light field (LF) and multi-view benchmark datasets demonstrate the significant superiority of our method over state-of-the-art methods. Especially, our proposed method employed to LF reconstruction task improves the PSNR of the conference version by around $1.0$ dB.

The rest of this paper is organized as follows. Section \ref{sec:related_work} reviews related
works. In Section \ref{sec:problem_analysis}, we analyze the drawbacks of the traditional image warping operation when used for view synthesis and propose content-aware warping. In Section \ref{sec:proposed_method}, we present the proposed view synthesis framework. In Section \ref{sec:experimental_results}, we conduct extensive experiments and analysis to evaluate our framework on both LF and multi-view datasets and discuss the limitation of our method. Finally, Section \ref{sec:conclusion} concludes this paper.

\section{Related Work}
\label{sec:related_work}
Based on the degree of the geometry constraint among views, we roughly divide  existing view synthesis methods into two categories: LF-based view synthesis and multi-view-based view synthesis.

\subsection{LF-based View Synthesis}
Existing LF reconstruction methods could be roughly divided into two categories: non-learning-based methods and learning-based methods.

Non-learning-based methods usually solve this inverse problem by regularizing th LF data based on different prior assumptions, e.g., Gaussian-based priors \cite{levin2008understanding, levin2010linear, mitra2012light}, sparse priors \cite{marwah2013compressive, shi2014light, vagharshakyan2017light}, and low-rank \cite{kamal2016tensor}. These methods either require many sparse samplings, or have high computational complexity. Another kind of methods for LF reconstruction is explicitly estimating the scene depth information, and then using it to warp input sub-aperture images (SAIs) to synthesize novel ones. Wanner and Goldluecke \cite{wanner2013variational} estimated disparity maps at input view by calculating the structure tensor of epipolar plane images (EPIs), and then used the estimated disparity maps to warp input SAIs to the novel viewpoints. This method makes the reconstruction quality rely heavily on the accuracy of the depth estimation. Zhang \textit{et al.} \cite{zhang2015light} proposed a disparity-assisted phase-based method that can iteratively refine the disparity map to minimize the phase difference between the warped novel SAI and the input SAI. However, the angular positions of synthesized SAIs are restricted to the neighbor of input views, which cannot reconstruct LFs from the input with large baselines.

Recently, many deep learning-based methods have been proposed to reconstruct dense LFs from sparse samplings. Yoon \textit{et al.} \cite{yoon2015learning} reconstructed novel SAIs from spatially up-sampled horizontal, vertical and surrounding SAI-pairs by using three separate networks. This method can only regress novel SAIs from adjacent ones, and could not process sparse LFs with large disparities. Wu \textit{et al.} \cite{wu2017light} used a 2-D image super-resolution network to recover high-frequency details along the angular dimension of the interpolated EPI. Analogously, Wang \textit{et al.} \cite{wang2018end} restored the high-frequency details of EPI stacks by using 3-D convolutional neural networks (CNNs). These methods process 2-D or 3-D slices of a 4-D sparse LF, which cannot fully explore the spatial-angular correlations implied in the LF. Yeung \textit{et al.} \cite{yeung2018fast} proposed an end-to-end network to reconstruct a dense LF from a sparse one in a single forward pass, which uses the spatial-angular separable convolution to explore the 4-D angular information effectively and efficiently. Meng \textit{et al.} \cite{meng2019high} directly employed the 4-D convolution to model the high-dimensional distribution of the LF data, and proposed a coarse-to-fine framework for spatial and angular super-resolution.
However, these regression-based methods always suffer from blurry effects or artifacts when input SAIs have relatively large baselines.

To handle sparse LFs with large baselines, 
some learning-based methods also employ
the pipeline of warping-based view synthesis. 
Kalantari \textit{et al.} \cite{kalantari2016learning} used two sequential networks to separately estimate the disparity map at the novel view, and predicted the color of novel SAI from warped images, respectively. Wu \textit{et al.} \cite{wu2019learning} extracted depth information from the sheared EPI volume, and then used it to reconstruct high angular-resolution EPIs. These methods either ignore the angular relations between synthesized SAIs, or underuse the spatial information of the input SAIs during the reconstruction. Srinivasan \textit{et al.} \cite{srinivasan2017learning} reconstructed an LF from a single 2-D image with predicting 4-D ray depths. This method only works on dataset with small disparities, and is restricted by its generalization ability. Jin \textit{et al.} \cite{jin2020deep} explicitly learned the disparity map at the novel viewpoint from input SAIs. They synthesized the coarse novel SAIs individually by fusing the warped input SAIs with confidence maps. Then they used a refinement network to recover the parallax structure by exploring the complementary information from the coarse LF. 

\subsection{Multi-view-based View Synthesis}

To synthesize novel views from a set of views, early IBR methods typically blend corresponding pixels from source views. These methods focus on improving the recovery of the scene geometry \cite{debevec1996modeling,chaurasia2013depth,hedman2016scalable} or modulating the blending weights \cite{levoy1996light, penner2017soft}, e.g., Penner and zhang \cite{penner2017soft} constructed a soft volume for each source view, where each voxel encodes a surface/free space confidence value, and then generated a novel view by performing back-to-front synthesis, based on the ray visibility and occlusion modeled from soft volumes. Recently, many deep learning-based methods also adopt the similar scheme and aim at improving the depth estimation, source-to-target blending, and post-processing refinement. Specifically, Hedman \textit{et al.} \cite{hedman2018deep} first generated per-source view 3D meshes by leveraging  advantages of two MVS reconstruction methods and using their proposed mesh simplification method. Then, they employed a CNN with four source-view mosaics and a rendered novel view image from the global mesh as inputs to predict blending weights, which are finally used to calculate the weighted sum of the five inputs as a novel view. Choi \textit{et al.} \cite{choi2019extreme} estimated the depth map of the novel view by warping and fusing source depth probability volumes, and used the depth map to synthesize an intermediate novel view by warping and blending. They also employed a patch-based refinement module to further recover the image details for the intermediate result.
Riegler and Koltun \cite{riegler2020free} obtained the novel view depth from the surface mesh. Based on the depth map, they warped the encoded source feature maps to the novel view, and then blended them via another network to further decode the novel view image. Shi \textit{et al.} \cite{shi2021self} warped source view feature maps to the target view to construct a feature frustum, and then compared their similarity to estimate the source visibility as well as the target depth probability, which are further used to warp and aggregate the source views to produce the novel view.

More recently, some learning-based methods for view synthesis propose to introduce specific scene representations with corresponding differentiable rendering procedures, e.g., MPI \cite{zhou2018stereo, flynn2019deepview, mildenhall2019local, srinivasan2019pushing, tucker2020single}, and NeRF \cite{mildenhall2020nerf,yu2021pixelnerf,wang2021ibrnet}. MPI is composed of a set of fronto-parallel planes at discrete depths, and each plane consists of an RGB image and an alpha image to encode the color and visibility information at the current depth.
The novel view can be rendered from the MPI by compositing the color images in back-to-front order using the differentiable over operation \cite{porter1984compositing}. Zhou \textit{et al.} \cite{zhou2018stereo} predicted the 
MPI at a reference view by using a CNN to represent the scene's content. Then the novel view can be synthesized from the MPI representation with homography and alpha compositing. Flynn \textit{et al.} \cite{flynn2019deepview} estimated an initial MPI from the source views, and then iteratively improved the initial MPI via learned updates which incorporate visibility information to improve the performance. Mildenhall \textit{et al.} \cite{mildenhall2019local} first independently expanded each source view to a local MPI, and then rendered the novel view by fusing adjacent local MPIs.
NeRF \cite{mildenhall2020nerf} use a multilayer perceptron (MLP) as an implicit function to represent the continuous radiance field, which outputs the view-dependent color and volume density at each 3-D spatial position. Then the novel view can be rendered through the differentiable and classical volume rendering \cite{kajiya1984ray}.
Afterwards, Yu \textit{et al.} \cite{yu2021pixelnerf} and Wang \textit{et al.} \cite{wang2021ibrnet} proposed to input image feature together with spatial coordinate and view direction to the MLP so as to learn generic NeRF functions that can generalize to other unseen scenes.

\begin{figure}[t]
\begin{center}
  \includegraphics[width=\linewidth]{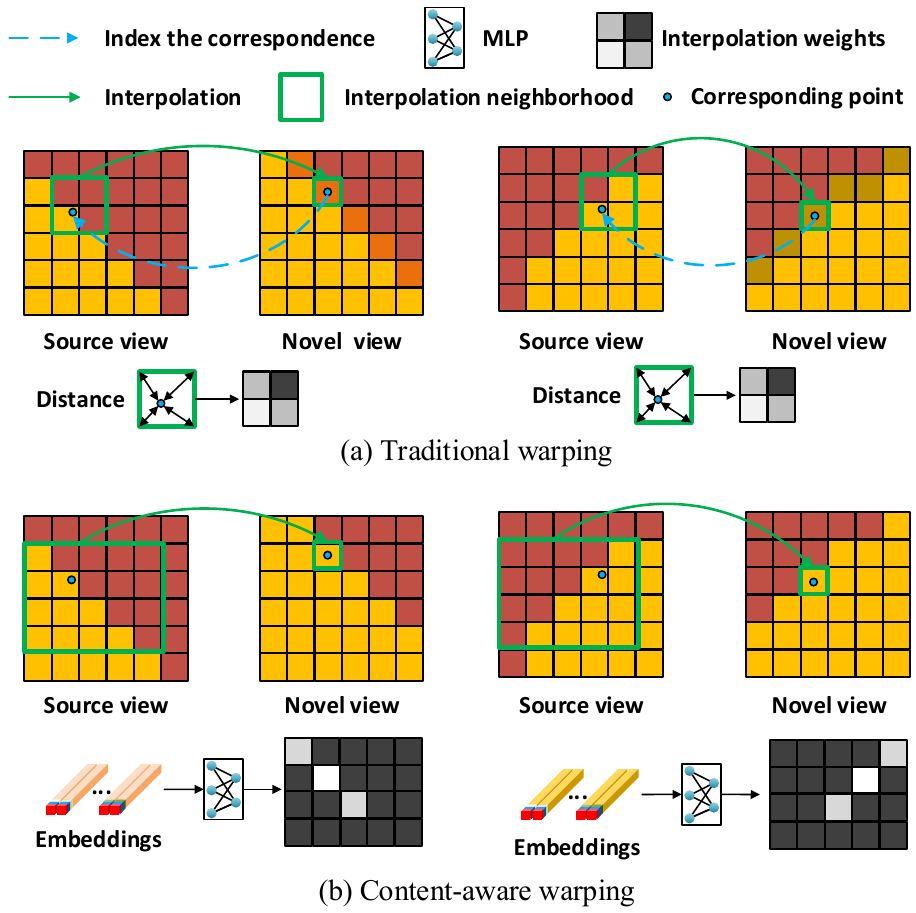}
\end{center}
\vspace{-0.3cm}
  \caption{Comparison of the traditional image warping operation and the proposed content-aware warping.
  In contrast to the content-independent weights employed in the warping operation
  (taking bilinear interpolation weights as an example),
  we propose to learn geometry-aware and content-adaptive interpolation weights from carefully constructed embeddings.
  }
\vspace{-0.4cm}
\label{fig:idea}
\end{figure}
\begin{figure*}[t]
\begin{center} 
  \includegraphics[width=\linewidth]{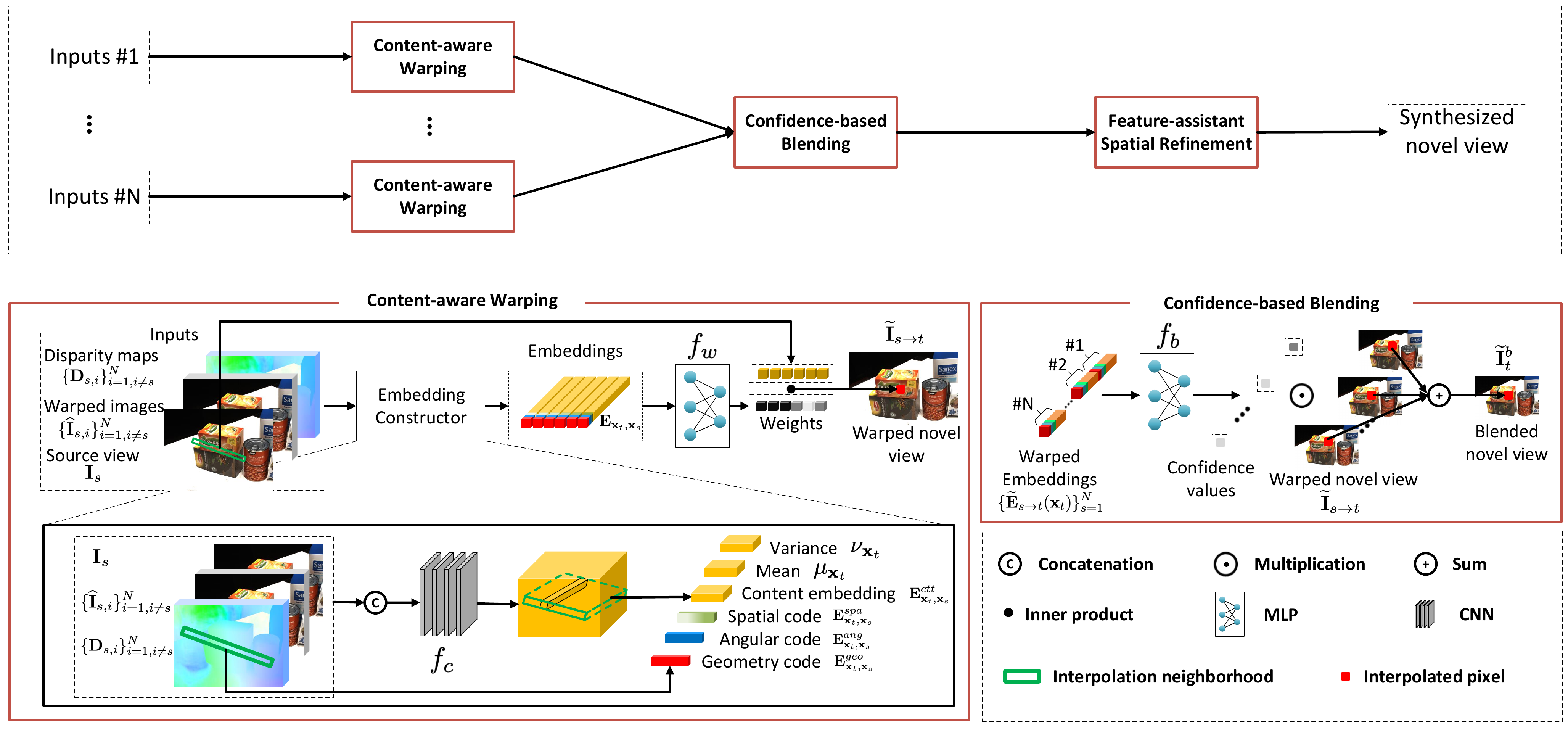}
\end{center}
\vspace{-0.2cm}
  \caption{Flowchart of the proposed framework
for view synthesis from $N$ input source views. It consists of three modules: content-aware warping, confidence-based blending, and feature-assistant spatial refinement. We refer the readers to Fig. \ref{fig:refinement} for more details of the feature-assistant spatial refinement module.
}
\vspace{-0.2cm}
\label{fig:pipeline}
\end{figure*}

\section{Analysis of Classic 2D Image Warping}
\label{sec:problem_analysis}

Given $N$ input source views denoted by $\{\mathbf{I}_s\}_{s=1}^{N}$, 
we aim to synthesize the unsampled novel view  between them, 
denoted by $\widetilde{\mathbf{I}}_t$, which should be as close to the ground-truth $\mathbf{I}_t$ as possible. 

Let $\mathbf{I}_{t}(\mathbf{x}_t)$ and $\mathbf{I}_{s}(\mathbf{x}'_t)$ be the projections of a typical scene point in different views, where $\mathbf{x}_t$ and $\mathbf{x}'_t$ are the 2D spatial coordinates of the pixels,
and under the assumption of Lambertian, we have
\begin{equation}
    \label{eq:LFstructure}
    \mathbf{I}_{t}(\mathbf{x}_t) = \mathbf{I}_{s}(\mathbf{x}'_t).
\end{equation}
Moreover, without occlusions, the relation between $\mathbf{x}_t$ and $\mathbf{x}'_t$ can be computed as\footnote{We do not distinguish between pixel coordinate and homogeneous coordinate for simplification.} 
\begin{equation}
\mathbf{x}'_t = \mathbf{K}_s(\mathbf{R}_s\mathbf{R}_t^\top d\mathbf{K}_t^{-1}\mathbf{x}_t+\mathbf{t}_s-\mathbf{R}_s\mathbf{R}_t^\top\mathbf{t}_t),
\label{equ:coordinate relation}
\end{equation}
where $d$ is the depth value of pixel $\mathbf{I}_{t}(\mathbf{x}_t)$,  and $\mathbf{K}_s$, $\mathbf{R}_s$, and $\mathbf{t}_s$ (resp. $\mathbf{K}_t$, $\mathbf{R}_t$, and $\mathbf{t}_t$) are the intrinsic matrix, the rotation matrix, and the translation vector of the source (resp. novel) view, respectively.
Thus, to synthesize $\widetilde{\mathbf{I}}_t$, for each pixel position of $\widetilde{\mathbf{I}}_t$, one can figure out the corresponding position in $\mathbf{I}_s$ according to Eq. \eqref{equ:coordinate relation}, and then map its pixel value to $\widetilde{\mathbf{I}}_t$. 
However, as the values of $\mathbf{x}'_t$ are not always integers, interpolation has to be performed to compute the intensity of the corresponding pixel, and the process can be formulated as
\begin{equation}
    \widetilde{\mathbf{I}}_{t}(\mathbf{x}_t) = \sum_{\mathbf{x}_s\in\mathcal{P}_{\mathbf{x}'_t}} w(\mathbf{x}_s-\mathbf{x}'_t ; \phi_w ) \mathbf{I}_s (\mathbf{x}_s),
\end{equation}
where $\mathcal{P}_{\mathbf{x}'_t}$ is the set of 2D coordinates of the pixels neighbouring to $\mathbf{x}'_t$, and function $w(\cdot; \cdot)$ with the parameter $\phi_w$ defines the interpolation weights for the pixels of $\mathbf{I}_s$, based on the distance between two pixels.

The above mentioned procedure is the image  backward warping operation widely-used in view synthesis. However, we argue that this procedure has natural limitations. Specifically,
as shown in Fig.~\ref{fig:idea} (a), 
this process 
pre-defines the neighbors used for synthesizing the target pixel based on the adopted  
interpolation kernel, e.g.,  $2\times 2$ pixels around the corresponding point are selected as the neighbors for bilinear interpolation \cite{jaderberg2015spatial}. Besides, the weight of each neighbor is only the function of distance from the corresponding point.
Thus, it is difficult to produce high-quality results, especially on areas with texture edges, occlusion boundaries, and non-Lambertian objects.

To overcome  the limitations of the traditional warping operation, as shown in Fig. \ref{fig:idea} (b), we propose \textit{content-aware warping}, which adaptively learns a weight value for each pixel of a relatively \textit{large} interpolation neighborhood, based on the contextual information. See Section \ref{subsec:weight learning} for the detailed process. We expect that such a process is able to assign a large weight 
to the neighbor, which is prone to the correspondence of the target pixel or semantically close to the target pixel, to emphasize its contribution, but a small one close to zero to the neighbor 
with a low probability of being the correspondence to exclude its interference. 
Based on this learnable warping process, we construct a new view synthesis framework explained in the next section.

Bako \textit{et al.} \cite{bako2017kernel} denoised the Monte Carlo rendering by separately filtering its diffusion and specular components. For each component, they fed the block of pre-processed color channel and auxiliary features centered at each pixel into a CNN to learn regular 2D kernels, which are further applied to the noisy color channel. Generally, we share the same motivation as Bako \textit{et al.} \cite{bako2017kernel}, i.e., overcoming the natural limitations of pre-defined and fixed kernels. However, as we focus on the problem of novel view synthesis,  the proposed method is technically very different from that in Bako \textit{et al.} \cite{bako2017kernel}.

\section{Proposed Method}
\label{sec:proposed_method}
\textbf{Overview}. As shown in Fig.~\ref{fig:pipeline}, the proposed view synthesis framework mainly consists of three modules, i.e., content-aware warping, confidence-based blending, and feature-assistant spatial refinement.
Specifically, given a set of input source views, 
we first warp them separately to the novel view 
pixel-by-pixel via content-aware warping (Section \ref{subsec:weight learning}), and then adaptively leverage the warped views
via confidence-based blending 
to handle the occlusion problem, leading to an intermediate result of the novel view (Section \ref{subsec:blending}). 
Finally, we explore the spatial correlation among the pixels of the intermediate result 
that are independently synthesized to further improve reconstruction quality via feature-assistant spatial refinement (Section \ref{subsec:refinement}), generating the final synthesized novel view. 
Besides, we also propose a weight-smoothness loss to regularize the network (Section \ref{subsec:loss}).
In what follows, we will introduce the technical details of each module. 

\subsection{Content-aware Warping }
\label{subsec:weight learning}
As aforementioned, in contrast to the traditional warping operation, the proposed content-aware warping adaptively learn the interpolation weight for each pixel of a relatively large neighborhood to synthesize the  pixel of the novel view. 
Note that we  warp the input source views separately to 
the novel view. 
Specifically, for each pixel of the novel view to be synthesized, we construct a neighborhood  $\mathcal{P}_{\mathbf{x}_t}$ by selecting pixels from  $\mathbf{I}_s$ based on the epipolar line corresponding to $\mathbf{x}_t$, i.e., $\mathcal{P}_{\mathbf{x}_t} = \{\mathbf{x}_s|\mathbf{x}_s\mathbf{F}\mathbf{x}_t=0\}$, where $\mathbf{F}$ is the fundamental matrix between the novel view and $\mathbf{I}_s$.
We then employ an MLP to predict the interpolation weight for each neighbor involved in the neighborhood by embedding the following information.
 
\textbf{(1) The correspondence relation between $\mathbf{I}_t$ and $\mathbf{I}_s$}. This cue helps to implicitly locate the corresponding pixel of $\mathbf{I}_t(\mathbf{x}_t)$ in $\mathbf{I}_s$.
The correspondence embedding consists of three components, i.e., a geometric code $\mathbf{E}_{\mathbf{x}_t,\mathbf{x}_s}^{geo}$, a spatial code $\mathbf{E}_{\mathbf{x}_t,\mathbf{x}_s}^{spa}$, and an angular code $\mathbf{E}_{\mathbf{x}_t,\mathbf{x}_s}^{ang}$.
Specifically, 
$\mathbf{E}_{\mathbf{x}_t,\mathbf{x}_s}^{geo}$ is the concatenation of disparity values of $\mathbf{I}_s(\mathbf{x}_s)$:
\begin{equation}
    \begin{aligned}
       \mathbf{E}_{\mathbf{x}_t,\mathbf{x}_s}^{geo} = \texttt{CAT}\left(\{\mathbf{D}_{s,i}(\mathbf{x}_s)\}_{i=1,i\ne s}^{N}\right),
    \end{aligned}
\end{equation}
where $\mathbf{D}_{s,i}$ is the disparity map of $\mathbf{I}_s$ calculated from $\mathbf{I}_s$ and another one of the remaining source views $\mathbf{I}_i$ by applying  an off-the-shelf disparity/optical flow estimation method. In this paper, we adopt the pre-trained RAFT \cite{teed2020raft} to estimate $\mathbf{D}_{s,i}$.
$\mathbf{E}_{\mathbf{x}_t,\mathbf{x}_s}^{spa}$ describes the spatial distance between $\mathbf{I}_t(\mathbf{x}_t)$ and $\mathbf{I}_s(\mathbf{x}_s)$, defined as 
\begin{equation}
    \begin{aligned}
        \mathbf{E}_{\mathbf{x}_t,\mathbf{x}_s}^{spa} &= \mathbf{x}_s - \mathbf{x}_t.
    \end{aligned}
\end{equation}
$\mathbf{E}_{\mathbf{x}_t,\mathbf{x}_s}^{ang}$ describes the relative camera pose between $\mathbf{I}_t$ and $\mathbf{I}_s$. A simple way to embed camera pose into the MLP is vectorizing the rotation and translation matrix and concatenating their entries with other codes, which dramatically increases the number of MLP parameters. To tackle this problem, 
we employ the 6DoF vector \cite{sun2018multi}  formed by 3-dimension translation vector and 3-dimension Euler angles to represent the camera pose.
We define $\mathbf{E}_{\mathbf{x}_t,\mathbf{x}_s}^{ang}$ as the difference of the 6DoF camera poses between $\mathbf{I}_t(\mathbf{x}_t)$ and $\mathbf{I}_s(\mathbf{x}_s)$, i.e.,
\begin{equation}
    \begin{aligned}
       \mathbf{E}_{\mathbf{x}_t,\mathbf{x}_s}^{ang} = \mathbf{z}_s - \mathbf{z}_t,
    \end{aligned}
\end{equation}
where $\mathbf{z}_s$ and $\mathbf{z}_t$ are the 6DoF camera poses of $\mathbf{I}_s$ and $\mathbf{I}_t$, respectively.

Given the disparities between $\mathbf{I}_s$ and other source views, we expect the network could implicitly infer the correspondence information between $\mathbf{I}_s$ and $\mathbf{I}_t$. Such correspondence relation between $\mathbf{I}_s$ and $\mathbf{I}_t$, together with the spatial code and angular code, would be an indicator for the MLP to determine whether the neighbor corresponds to the target pixel $\mathbf{I}_t(\mathbf{x}_t)$, and assigns appropriate weight values to the neighbor. 

\begin{figure*}[thp]
\begin{center}
  \includegraphics[width=\linewidth]{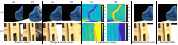}
\end{center}
\vspace{-0.4cm}
  \caption{Visual illustration of the effect of the confidence-based blending.}
\vspace{-0.3cm}
\label{fig:visual_conf}
\end{figure*}

\textbf{(2) The content information around $\mathbf{I}_s(\mathbf{x}_s)$}. This kind of information helps to understand complicated scenarios, such as texture edges, occlusion boundaries, and non-Lambertian objects.
To construct the content embedding, denoted by $\mathbf{E}_{\mathbf{x}_t,\mathbf{x}_s}^{ctt}$,
we first  separately backward warp the remaining input views $\{\mathbf{I}_i\}_{i=1,i\ne s}^{N}$ to $\mathbf{I}_s$, based on $\{\mathbf{D}_{s,i}\}_{i=1,i\ne s}^{N}$, generating $\{\widehat{\mathbf{I}}_{s,i}\}_{i=1,i\ne s}^{N}$.
Then, we employ a sub-CNN $f_c(\cdot)$ to learn the content information, i.e.,
\begin{equation}
    \begin{aligned}
        \mathbf{E}_{\mathbf{x}_t,\mathbf{x}_s}^{ctt} &= f_c\left(\mathbf{x}_s, \mathbf{x}_t,\mathbf{I}_s, \{\widehat{\mathbf{I}}_{s,i}\}_{i=1,i\ne s}^{N}, \{\mathbf{D}_{s,i}\}_{i=1,i\ne s}^{N}\right).
    \end{aligned}
\end{equation}
It is expected that $f_c(\cdot)$ is able to detect the texture edges of $\mathbf{I}_s$ and  
understand the occlusion and non-Lambertian relations by comparing $\mathbf{I}_s$ and $\{\widehat{\mathbf{I}}_{s,i}\}_{i=1,i\ne s}^{N}$ with the assistance of $\{\mathbf{D}_{s,i}\}_{i=1,i\ne s}^{N}$. 
In spite of per-pixel content information, we explicitly calculate the global mean $\mathbf{\mu}_{\mathbf{x}_t}$ and variance $\mathbf{\nu}_{\mathbf{x}_t}$ of $\mathbf{E}_{\mathbf{x}_t,\mathbf{x}_s}^{ctt}$ over the neighborhood $\mathcal{P}_{\mathbf{x}_t}$: 
\begin{equation}
    \begin{aligned}
        &\mathbf{\mu}_{\mathbf{x}_t} = \frac{1}{|\mathcal{P}_{\mathbf{x}_t}|}\sum_{\mathbf{x}_s\in\mathcal{P}_{\mathbf{x}_t}} \mathbf{E}_{\mathbf{x}_t,\mathbf{x}_s}^{ctt},\\
        &\mathbf{\nu}_{\mathbf{x}_t} = \frac{1}{|\mathcal{P}_{\mathbf{x}_t}|}\sum_{\mathbf{x}_s\in\mathcal{P}_{\mathbf{x}_t}} (\mathbf{E}_{\mathbf{x}_t,\mathbf{x}_s}^{ctt}-\mathbf{\mu}_{\mathbf{x}_t})^2,
    \end{aligned}
\end{equation}
where $|\mathcal{P}_{\mathbf{x}_t}|$ is the number of pixels in $\mathcal{P}_{\mathbf{x}_t}$. 
The global mean and variance can help the MLP to distinguish which neighborhoods a typical source pixel belongs to, and assign the source pixel a
appropriate weight to improve the reconstruction quality.
We finally construct the geometry and content embedding
$\mathbf{E}_{\mathbf{x}_t,\mathbf{x}_s}$
as
\begin{equation}
    \begin{aligned}
        \mathbf{E}_{\mathbf{x}_t,\mathbf{x}_s} = \texttt{CAT}\left(
        \mathbf{E}_{\mathbf{x}_t,\mathbf{x}_s}^{geo}, \mathbf{E}_{\mathbf{x}_t,\mathbf{x}_s}^{spa},
        \mathbf{E}_{\mathbf{x}_t,\mathbf{x}_s}^{ang},
        \mathbf{E}_{\mathbf{x}_t,\mathbf{x}_s}^{ctt},
        \mathbf{\mu}_{\mathbf{x}_t},
        \mathbf{\nu}_{\mathbf{x}_t}
        \right),
    \end{aligned}
\end{equation}
where $\texttt{CAT}(\cdot)$ is the concatenation operation,
and predict the interpolation weights  $W_{\mathbf{x}_t,\mathbf{x}_s}$ 
as
\begin{equation}
    \begin{aligned}
        W_{\mathbf{x}_t,\mathbf{x}_s} = f_w\left(\mathbf{E}_{\mathbf{x}_t,\mathbf{x}_s}\right),
    \end{aligned}
\end{equation}
where $f_w(\cdot)$ is the learnable MLP.
We apply a softmax function to the outputs of the network $f_w(\cdot)$ over $\mathcal{P}_{\mathbf{x}_t}$ to ensure that the interpolated pixel color lies in the convex hull of the corresponding neighborhood in the source view.
With the learned weights, we can obtain the pixel of the warped view located at $\mathbf{x}_t$ as 
\begin{equation}
    \begin{aligned}
        \widetilde{\mathbf{I}}_{s\rightarrow t} (\mathbf{x}_t) = \sum_{\mathbf{x}_s\in\mathcal{P}_{\mathbf{x}_t}} W_{\mathbf{x}_t,\mathbf{x}_s} \mathbf{I}_s(\mathbf{x}_s).
    \end{aligned}
\end{equation}

\subsection{Confidence-based Blending} 
\label{subsec:blending}

Although the 
content-aware warping module has the ability of handling occlusion boundaries by embedding the contextual information,
it is still difficult to synthesize the pixels whose correspondences are completely occluded in a source view 
by only warping that source view. 
Considering that the object occluded from one viewpoint might be visible from other ones,
we thus blend the views separately warped 
from $\{\mathbf{I}_s\}_{s=1}^{N}$   
under the guidance of their confidence maps, which indicate the non-occlusion pixels with higher values.

To predict the confidence value for each pixel position $\mathbf{x}_t$ in the synthesized view, 
we first aggregate the geometry and content embeddings of all neighbors corresponding to $\mathbf{x}_t$ in $\mathbf{I}_s$ by the content-aware warping, i.e., 
\begin{equation}
    \begin{aligned}
        \widetilde{\mathbf{E}}_{s\rightarrow t} (\mathbf{x}_t) = \sum_{\mathbf{x}_s\in\mathcal{P}_{\mathbf{x}_t}} W_{\mathbf{x}_t,\mathbf{x}_s} \mathbf{E}_{\mathbf{x}_t,\mathbf{x}_s},
    \end{aligned}
\end{equation}
where $\widetilde{\mathbf{E}}_{s\rightarrow t} (\mathbf{x}_t)$ is the aggregated content and geometry embedding corresponding to $\mathbf{x}_t$ from $\mathbf{I}_s$.
We then apply another MLP, denoted by $f_b(\cdot)$, on the concatenation of the aggregated embeddings from all source views to predict confidence values corresponding to them, i.e.,
\begin{equation}
\begin{aligned}
\widetilde{\mathcal{C}}_{t}(\mathbf{x_t}) = f_b\left(\texttt{CAT}\left(\{\widetilde{\mathbf{E}}_{s\rightarrow t}(\mathbf{x}_t)\}_{s=1}^{N}\right)\right),
\end{aligned}
\end{equation}
where $\widetilde{\mathcal{C}}_{t}\in \mathbb{R}^{N\times H\times W}$ is the confidence map volume formed by the predicted confidence maps for $\{\widetilde{\mathbf{I}}_{s\rightarrow t}\}_{s=1}^{N}$, and let $\widetilde{\mathbf{C}}_{t}^{s}\in \mathbb{R}^{H\times W}$ be the $s$-th slice, i.e., the confidence map for $\widetilde{\mathbf{I}}_{s\rightarrow t}$, where $H$ and $W$ are the spatial dimensions of the input source views.
Based on the learned confidence maps, we then blend the warped views to   
produce the intermediate 
result of the novel view as
\begin{equation}
\begin{aligned}
\widetilde{\mathbf{I}}_t^b = \sum_{s=1}^{N} \widetilde{\mathbf{C}}_{t}^{s}\odot \widetilde{\mathbf{I}}_{s\rightarrow t},
\end{aligned}
\end{equation}
where $\odot$ is the element-wise multiplication operator.

As an example,  Fig.~\ref{fig:visual_conf} visually illustrates the advantage of such a confidence-based blending module.  

\begin{figure*}[t]
\begin{center} 
  \includegraphics[width=\linewidth]{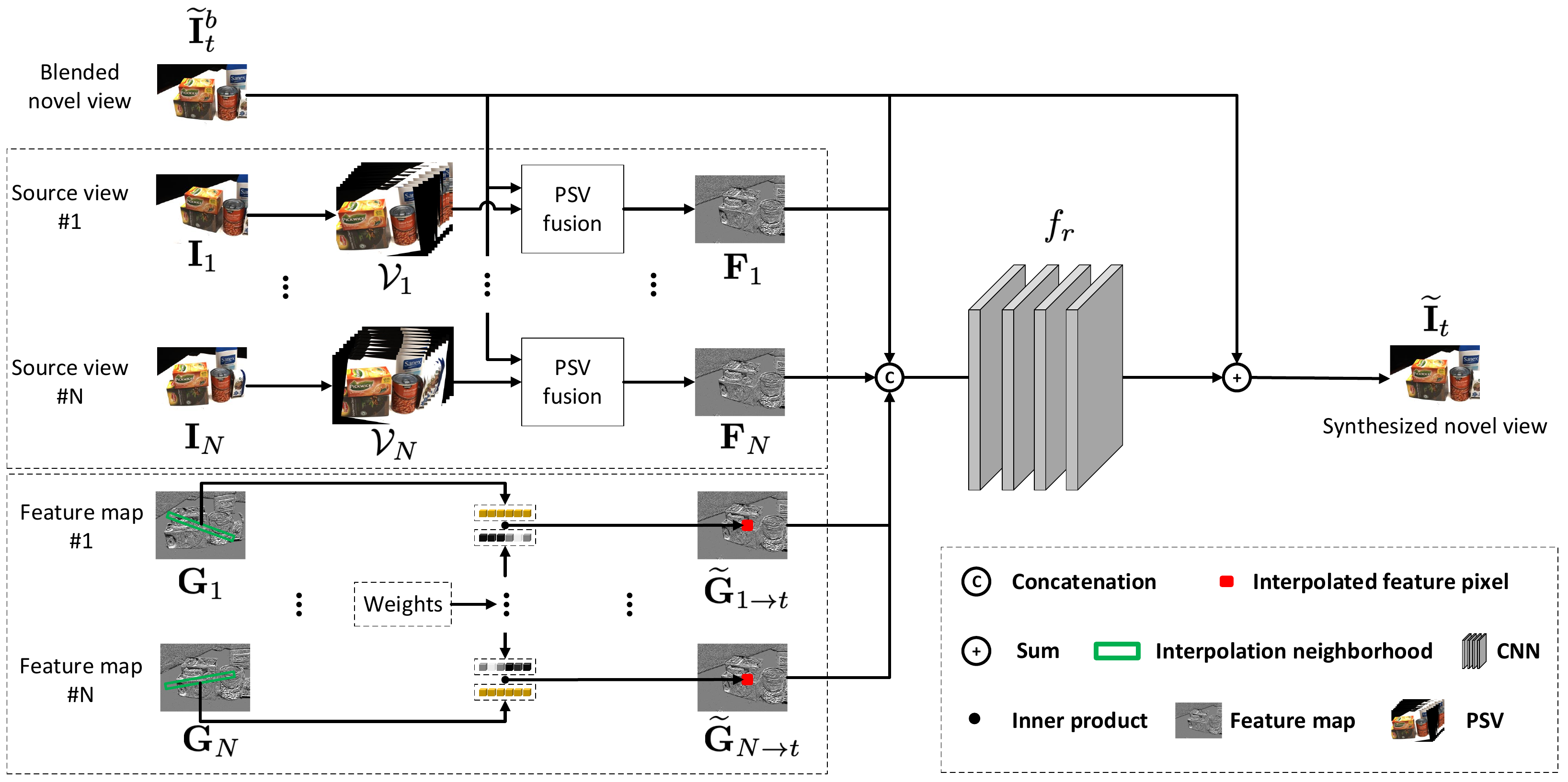}
\end{center}
  \caption{
Flowchart of the proposed feature-assistant spatial refinement module.
}
\vspace{-0.2cm}
\label{fig:refinement}
\end{figure*}

\subsection{Feature-assistant Spatial Refinement}
\label{subsec:refinement}
As the pixels of the intermediate result $\widetilde{\mathbf{I}}_t^b$ are independently synthesized, the spatial correlation among them is not taken into account, i.e., the intensities of pixels that are spatially close are generally similar.
Thus, as shown in Fig.~\ref{fig:refinement}, we propose to propagate the spatial correlation in $\mathbf{I}_s$ to the corresponding region of  $\widetilde{\mathbf{I}}_t^b$ in both image and feature space to further refine the quality of $\widetilde{\mathbf{I}}_t^b$.
Generally, in image space, 
we warp $\mathbf{I}_s$ to the target view via constructing a plane-sweep volume (PSV) \cite{collins1996space}, then adaptively aggregates useful information from
the warped source images at different depth layers in the constructed PSV.
In feature space, we align the feature map extracted from $\mathbf{I}_s$ to the novel view via the proposed content-aware warping. The feature-space warping is based on the observation that a typical pixel of the source view mapped to the feature space encodes the information of its local region containing the spatial correlation around it.
We learn an additive map for $\widetilde{\mathbf{I}}_t^b$ by using a sub-CNN with the concatenation of the following information: $\widetilde{\mathbf{I}}_t^b$, the aggregated PSV feature maps and the content-aware-warping aligned feature maps from $\{\mathbf{I}_s\}_{s=1}^{N}$.

Specifically, let $\mathcal{V}_s\in \mathbb{R}^{D\times H\times W\times C}$ be the PSV constructed from $\mathbf{I}_s$ with uniformly sampling $D$ depth layers within the depth range of the target view, and $\mathbf{V}_s^d\in \mathbb{R}^{H\times W\times C}$ the warped image at $d$-th ($1\leq d \leq D$) depth layer, where $C$ is the channel number of the source views. To adaptively aggregate the information across depth layers, we extract a feature map as well as predict a corresponding weight value at a typical depth layer $d$ by employing a sub-CNN $f_v(\cdot)$ with the concatenation of $\widetilde{\mathbf{I}}_t^b$ and $\mathbf{V}_s^d$ as input, i.e., 
\begin{equation}
    \begin{aligned}
        \mathbf{F}_s^d, w_s^d = f_v\left(\texttt{CAT}\left(\widetilde{\mathbf{I}}_t^b, \mathbf{V}_s^d\right)\right),
    \end{aligned}
\end{equation}
where $\mathbf{F}_s^d\in \mathbb{R}^{H\times W\times 64}$ is the feature map corresponding to $\mathbf{V}_s^d$, and $w_s^d$ is the weight value. $f_v(\cdot)$ estimates a $65$-channel output where first $64$-channel is $\mathbf{F}_s^d$ and the last channel is for the weight. We average the values of the last channel as the weight value $ w_s^d$, and
normalize the weight values from different depth layers by a softmax function. We then fuse the feature maps from different depth layers as
\begin{equation}
    \begin{aligned}
        \mathbf{F}_s = \sum_{d=1}^D w_s^d \mathbf{F}_s^d,
    \end{aligned}
\end{equation}
where $\mathbf{F}_s$ is the fused PSV feature map from $\mathbf{I}_s$.

Furthermore, we map $\mathbf{I}_s$ to the feature space by employing a sub-CNN $f_g(\cdot)$, i.e.,
\begin{equation}
    \begin{aligned}
        \mathbf{G}_s = f_g(\mathbf{I}_s),
    \end{aligned}
\end{equation}
where $\mathbf{G}_s$ is the feature map of $\mathbf{I}_s$. Then, we align $\mathbf{G}_s$ with the feature map of the novel view via the content-aware warping with the learned weights in Section \ref{subsec:weight learning} to perceive the corresponding information from $\mathbf{G}_s$, i.e.,
\begin{equation}
    \begin{aligned}
        \widetilde{\mathbf{G}}_{s\rightarrow t} (\mathbf{x}_t) = \sum_{\mathbf{x}_s\in\mathcal{P}_{\mathbf{x}_t}} W_{\mathbf{x}_t,\mathbf{x}_s} \mathbf{G}_s(\mathbf{x}_s).
    \end{aligned}
\end{equation}
We finally refine $\widetilde{\mathbf{I}}_t^b$ as
\begin{equation}
    \begin{aligned}
       \widetilde{\mathbf{I}}_t = f_r\left(\texttt{CAT}\left(\widetilde{\mathbf{I}}_t^b, \{\mathbf{F}_s\}_{s=1}^{N}, \{\mathbf{G}_s\}_{s=1}^{N} \right)\right) + \widetilde{\mathbf{I}}_t^b,
    \end{aligned}
\end{equation}
where $f_r(\cdot)$ denotes a sub-CNN.

\subsection{Loss Function}
\label{subsec:loss}
To train the network, basically we use the ground-truth novel view $\mathbf{I}_t$ to supervise both the final and intermediate predictions of the novel view, i.e., 
\begin{equation}
\begin{aligned}
        \ell^r_t = \left\|\widetilde{\mathbf{I}}_t-\mathbf{I}_t\right\|_1 
        + \left\|\widetilde{\mathbf{I}}_t^b-\mathbf{I}_t\right\|_1
        + \sum_{s=1}^{N} \left\| \widetilde{\mathbf{I}}_{s\rightarrow t}-\mathbf{I}_t\right\|_1,
\end{aligned}
\end{equation}
where $\|\cdot\|_1$ denotes the $L_1$ norm. 
Besides, we also employ the perceptual loss and ssim loss to further regularize the final prediction, which are defined as
\begin{equation}
\begin{aligned}
       \ell^p_t = 
        \left\|\phi(\widetilde{\mathbf{I}}_t)-\phi(\mathbf{I}_t)\right\|_1,
\end{aligned}
\end{equation}
and 
\begin{equation}
\begin{aligned}
        \ell^s_t = 
        \frac{1-SSIM(\widetilde{\mathbf{I}}_t,\mathbf{I}_t)}{2},
\end{aligned}
\end{equation}
respectively, where $\phi(\cdot)$ is the output of the layer `conv1\_2` of a pretrained VGG-19 network \cite{simonyan2015very}, and $SSIM(\cdot,\cdot)$ is the structural similarity index measure \cite{wang2004image} between two images.

Moreover, spatially close pixels generally tend to have similar intensity, and their neighborhoods used for interpolation are mostly overlapping. Thus, the distributions of learned weights should be similar. Based on this observation, we propose a weight-smoothness loss term to regularize the network. 
Specifically, let $\mathcal{W}_s \in \mathbb{R}^{|\mathcal{P}_{\mathbf{x}_t}|\times H\times W}$ be the weight map volume formed by the learned weights from $\mathbf{I}_s$ for synthesizing all pixels of $\widetilde{\mathbf{I}}_t$, and $\mathbf{W}_s^l \in \mathbb{R}^{ H\times W}$ the $l$-th ($1\leq l \leq |\mathcal{P}_{\mathbf{x}_t}|$) slice.  
Accordingly, the above-mentioned observation indicates that $\mathbf{W}_s^l$ should be locally constant. Therefore, we mathematically formulate the weight-smoothness loss as 
\begin{equation}
\begin{aligned}
\ell^w = \sum_{s=1}^{N}\sum_{l=1}^{|\mathcal{P}_{\mathbf{x}_t}|}\left(\left\|\nabla_x \mathbf{W}_s^l\right\|_1 +\left\|\nabla_y \mathbf{W}_s^l\right\|_1\right),
\end{aligned}
\end{equation}
where
$\nabla_x$ and $\nabla_y$ are the gradient operators for the spatial domain. 

In all, we  train the proposed framework end-to-end with the following loss function: 
\begin{equation}
\ell = \ell_t^r  + \ell_t^p + \ell_t^s + \lambda \ell^w,    
\end{equation}
where $\lambda\geq 0$ is the hyper-parameter to balance the two terms. 

\textit{Remark}. Compared with the disparity-oriented loss of the preliminary work \cite{guo2021deep}, the weight-smoothness loss is more practical because 
the disparity-oriented loss requires ground-truth disparity maps, 
which are unavailable for 
real-world LF data, 
while the weight-smoothness loss does not require these data. 

\begin{table*}[thp]
\centering
\setlength{\tabcolsep}{0.5mm}
\caption{Quantitative comparisons (PSNR/SSIM) of different methods on the Inria Sparse LF dataset \cite{shi2019framework}. The best and second best results are highlighted in \textcolor{red}{red} and \textcolor{blue}{blue}, respectively.}
\begin{tabular}{c|c|ccccccccc}
\toprule
        Light Field &\makecell{Disparity\\range} &\makecell{Baseline\\(Warp)} &\makecell{Baseline\\(Disparity)} & \makecell{Kalantari\\\textit{et al.} \cite{kalantari2016learning}}  & \makecell{Wu\\\textit{et al.} \cite{wu2019learning}} & \makecell{Wu\\\textit{et al.} \cite{wu2021revisiting}} & \makecell{Jin\\\textit{et al.} \cite{jin2020deep}} & \makecell{Guo\\\textit{et al.} \cite{guo2021learning}} & \makecell{Bao\\\textit{et al.} \cite{bao2019depth}} & \makecell{Ours}   \\ 
\midrule
Electro\_devices    &[-19.6, 32.8] &32.87/0.943 &32.99/0.941  &24.58/0.691 &29.12/0.865 &31.22/0.883 &33.04/0.935 &\textcolor{blue}{35.38}/\textcolor{blue}{0.959} &32.81/0.929  &\textcolor{red}{36.25}/\textcolor{red}{0.963}\\
Flying\_furniture   &[-34.0, 62.4] &31.04/0.901 &32.24/0.896  &28.99/0.786 &27.31/0.795 &31.19/0.858 &31.35/0.892 &\textcolor{blue}{33.64}/\textcolor{blue}{0.932} &31.51/0.892  &\textcolor{red}{35.71}/\textcolor{red}{0.947}\\
Coffee\_beans\_vases&[10.8, 58.4]  &27.43/0.920 &29.13/0.934  &21.56/0.582 &25.42/0.890 &26.57/0.856 &28.22/0.928 &29.37/\textcolor{blue}{0.942} &\textcolor{blue}{29.55}/0.932  &\textcolor{red}{30.48}/\textcolor{red}{0.950}\\
Dinosaur            &[-57.6, 72.8] &25.45/0.865 &27.20/0.896  &22.40/0.730 &24.21/0.819 &26.14/0.881 &\textcolor{blue}{27.23}/0.900 &27.19/\textcolor{blue}{0.902} &26.71/0.889  &\textcolor{red}{28.41}/\textcolor{red}{0.924}\\
Flowers             &[-40.4, 66.0] &23.88/0.806 &24.41/0.822  &22.02/0.669 &23.45/0.779 &23.73/0.817 &24.35/0.839 &24.80/0.845 &\textcolor{red}{26.74}/\textcolor{red}{0.878}  &\textcolor{blue}{25.13}/\textcolor{blue}{0.868}\\
Rooster\_clock      &[-34.4, 21.2] &36.46/0.952 &35.77/0.953  &22.73/0.710 &29.41/0.889 &25.15/0.884 &27.58/0.928 &\textcolor{red}{38.48}/\textcolor{blue}{0.966} &35.38/0.942  &\textcolor{blue}{38.26}/\textcolor{red}{0.968}\\
Smiling\_crowd      &[-40.4, 64.8] &20.75/0.822 &21.29/0.818  &17.01/0.602 &20.33/0.756 &20.36/0.777 &21.01/0.819 &\textcolor{blue}{22.34}/\textcolor{blue}{0.868} &20.86/0.816  &\textcolor{red}{22.79}/\textcolor{red}{0.877}\\
\bottomrule
\multicolumn{2}{c|}{\textbf{Average}} &28.27/0.887 &29.00/0.894 &22.76/0.682 &25.61/0.827 &26.34/0.851 &27.54/0.891 &\textcolor{blue}{30.17}/\textcolor{blue}{0.916} &29.08/0.897 &\textcolor{red}{31.00}/\textcolor{red}{0.928}\\
\bottomrule
\end{tabular}
\label{table:quantitative}
\end{table*}

\section{Experiments}
\label{sec:experimental_results}
\subsection{Implementation Details and Datasets}
The content embedding network $f_c(\cdot)$ and the spatial refinement network $f_r(\cdot)$ are all 2-D CNNs composed of residual blocks \cite{he2016deep} with the kernel of size $3\times3$. We utilized zeros-padding to keep the spatial size unchanged. The context extraction network $f_g(\cdot)$ is a U-Net architecture which is the same as the image encoding  CNN of \cite{riegler2021stable}.
We refer readers to the \textit{Supplementary Material} for the detailed network architecture. 
At each iteration of the training phase, we synthesized a fixed-size patch randomly cropped from the target image. 
The batch size was set to $4$. The learning rate was initially set to $1e^{-4}$  and reduced to $1e^{-5}$ after $8000$ epochs. We used Adam \cite{kingma2015adam} with $\beta_1=0.9$ and $\beta_2=0.999$ as the optimizer. 

We trained and tested our network on both LF and multi-view datasets. 
 For LF datasets, we reconstructed the 3-D LF containing $5$ SAIs, i.e., inputting SAIs at two ends as source views to reconstruct middle three ones.
Specifically, we trained our network with $29$ LF images from the Inria Sparse LF dataset \cite{shi2019framework}. Each LF image is 4-D and contains $9\times 9$ SAIs with a disparity range of $[-20,20]$ between adjacent SAIs. 
We extracted 3-D LFs as training samples from the 4-D LF image by taking the 3$^{rd}$ to 7$^{th}$ SAIs at each row. Note that the disparity between two input source views at each 3-D LF is up to $80$ pixels. 
The test dataset consists of $7$ LF images from the Inria Sparse LF dataset  \cite{shi2019framework}. For each LF image, we took the 3$^{rd}$ and 7$^{th}$ SAIs at the 5$^{th}$ row as input source views to reconstruct the 3-D LF.
We also tested on $14$ LF images from the MPI LF archive \cite{Vamsi2017}. Note that MPI \cite{Vamsi2017} is a high angular-resolution LF dataset where each LF image contains $101$ SAIs distributed on a scanline. Thus, we can construct testing LFs with different disparities by sampling SAIs with different intervals (see details in Section \ref{sec:lf_eval}).
For multi-view datasets, we trained and tested our network on both DTU\cite{aanaes2016large} and RealEstate10K\cite{zhou2018stereo} datasets. For DTU\cite{aanaes2016large} dataset, we used the dataset preprocessed by \cite{yao2018mvsnet}, and trained our network on $79$ scenes, and tested on $18$ scenes. For RealEstate10K\cite{zhou2018stereo} dataset, we trained our network on $85$ scenes, and tested on $17$ scenes. Given multiple views of a scene, we constructed a training sample by first randomly selecting a novel view, and then sampling its two adjacent ones as source views. 
Since different datasets have different disparity ranges, we set various interpolation neighborhood sizes in the source view for different datasets. We refer readers to the \textit{Supplementary Material} for the detailed construction of the interpolation neighborhood.

\subsection{Evaluation on LF Datasets}
\label{sec:lf_eval}
We compared the proposed method with five state-of-the-art deep learning-based LF reconstruction methods, including Kalantari \textit{et al.} \cite{kalantari2016learning}, Wu \textit{et al.} \cite{wu2019learning}, Wu \textit{et al.} \cite{wu2021revisiting}, Jin \textit{et al.} \cite{jin2020deep}\footnote{Note that the 2-D angular convolutional layers 
were degenerated to 1-D convolutional layers to adapt to the 3-D LFs.}, and Guo \textit{et al.} \cite{guo2021learning}. Besides, we also compared with a warping-based video frame interpolation method, i.e., Bao \textit{et al.} \cite{bao2019depth} which also employs a learned warping layer.
For a fair comparison, we trained all the methods  on the same dataset with the officially released codes and suggested configurations.
Since the disparity between source views, i.e., source-to-source disparity, and the disparity between the source and target views, i.e., source-to-target disparity, are proportional in the LF, we can easily calculate the source-to-target disparity from the source-to-source disparity. To let the MLP directly perceive correspondence relation between the source view and the target view, we used the source-to-target disparity instead of the source-to-source disparity to construct the geometry code and learn the content information in our method on the LF datasets.

To directly verify the advantage of the proposed content-aware warping over the traditional warping operation,
we constructed a baseline model, named Baseline (Warp), by replacing the content-aware warping module of our framework with a disparity-based warping operation while leaving other modules unchanged.
Specifically,  Baseline (Warp) first forward warps the disparity maps $\mathbf{D}_{1,2}$ and $\mathbf{D}_{2,1}$ to the novel view, and then employs a sub-CNN to predict the disparity map of the novel view and two confidence maps corresponding to the two input views from the concatenation of the warped disparity maps. Baseline (Warp) further backward warps the two input SAIs to the novel view separately based on the predicted disparity map, and blends them based on the confidence maps. Finally, the blended SAI is refined by the feature-assistant spatial refinement module. 

 \begin{figure}[thp]
\centering
\subfloat{\includegraphics[width=0.255\textwidth]{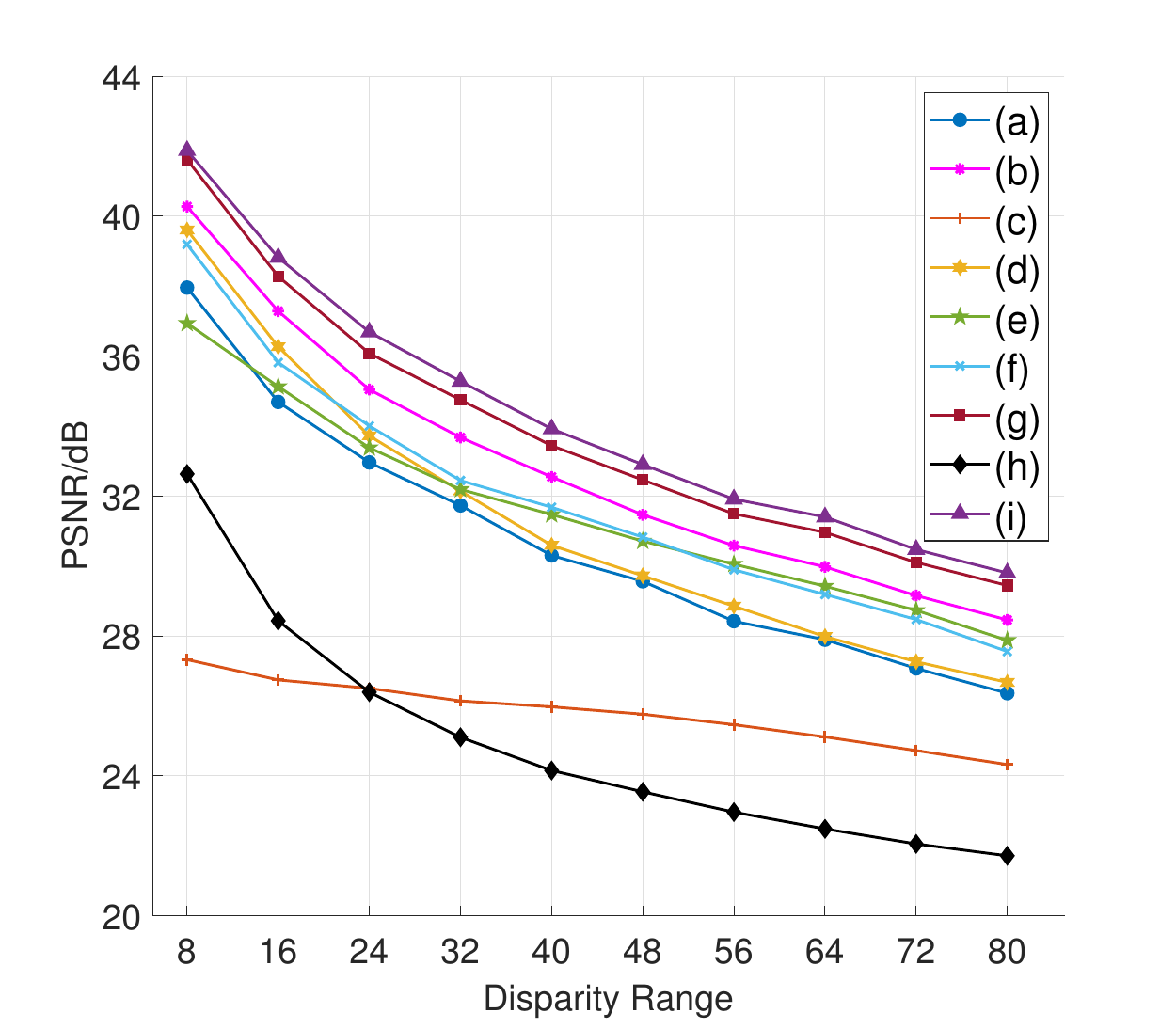}}
\subfloat{\includegraphics[width=0.255\textwidth]{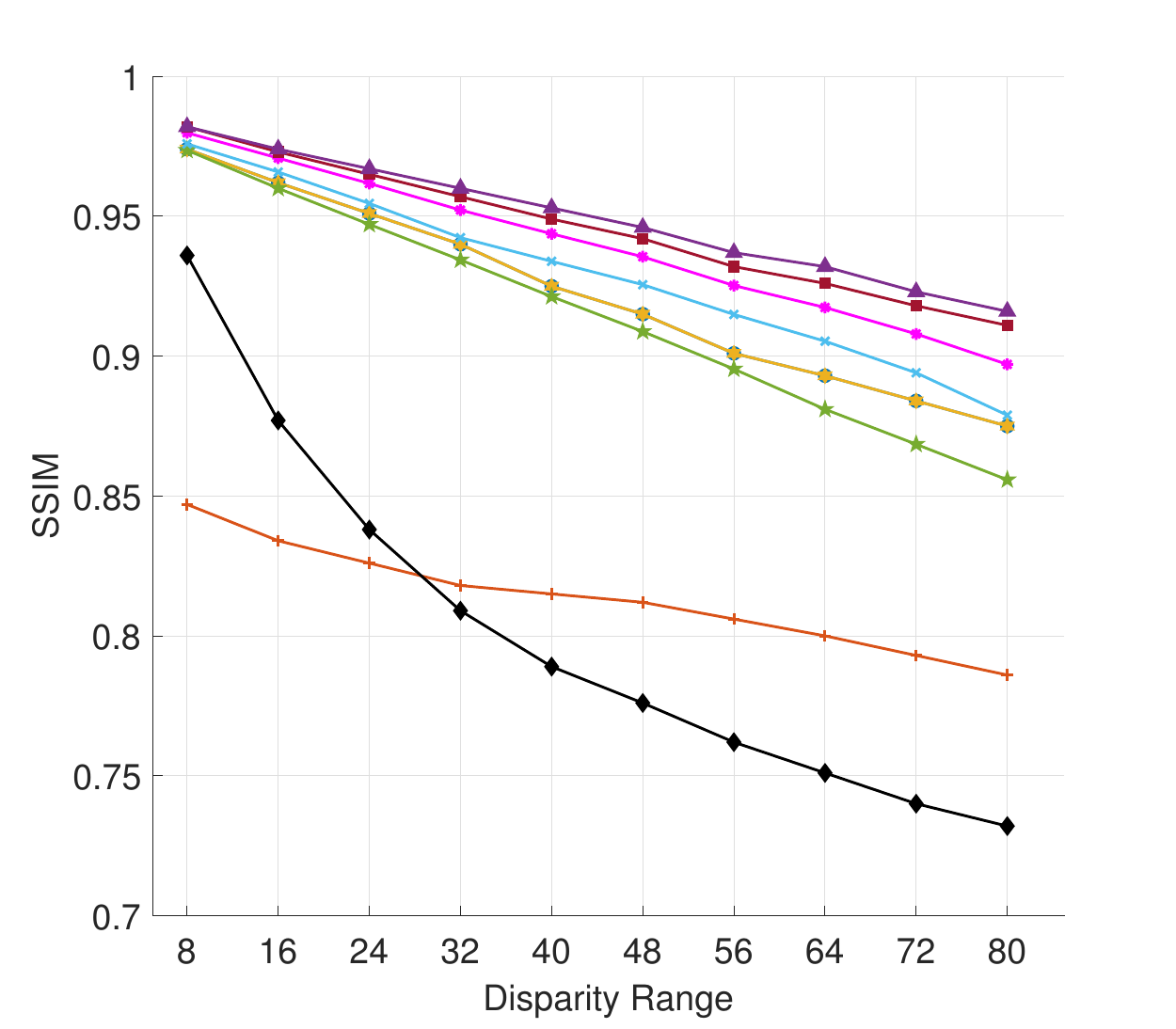}}
\caption{Quantitative comparisons (PSNR/SSIM) of different methods under various  disparity ranges (pixels) between input SAIs on  the MPI LF dataset \cite{Vamsi2017}. (a) Baseline (Warp), (b) Baseline (Disparity), (c) Kalantari \textit{et al.} \cite{kalantari2016learning}, (d) Wu \textit{et al.} \cite{wu2019learning}, (e) Wu \textit{et al.} \cite{wu2021revisiting}, (f) Jin \textit{et al.} \cite{jin2020deep}, (g) Guo \textit{et al.} \cite{guo2021learning}, (h) Bao \textit{et al.} \cite{bao2019depth}, (i) Ours. The two subfigures share the same legend.}
\vspace{-0.4cm}
\label{fig:quanti_mpi}
\end{figure}

\begin{figure*}[t]
\begin{center}
  \includegraphics[width=\linewidth]{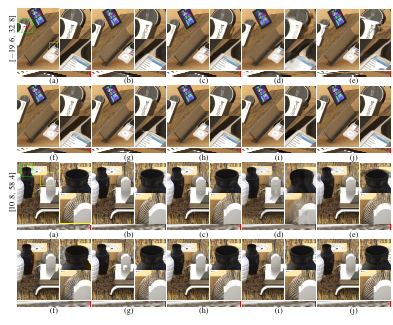}
\end{center}
\vspace{-0.4cm}
  \caption{Visual comparisons of reconstructed SAIs from different methods on the Inria Sparse LF dataset \cite{shi2019framework}. 
  (a) Ground Truth, (b) Baseline (Warp), (c) Baseline (Disparity), (d) Kalantari \textit{et al.} \cite{kalantari2016learning}, (e) Wu \textit{et al.} \cite{wu2019learning}, (f) Wu \textit{et al.} \cite{wu2021revisiting}, (g) Jin \textit{et al.} \cite{jin2020deep}, (h) Guo \textit{et al.} \cite{guo2021learning}, (i) Bao \textit{et al.} \cite{bao2019depth}, and (j) Ours. The disparity range between input SAIs of each LF is shown on the left.} 
\label{fig:visual_inria}
\end{figure*}

Besides, we also set another baseline, named Baseline (Disparity), by incorporating the estimated disparity maps used in our framework into Jin \textit{et al.}  \cite{jin2020deep}\footnote{Here we selected Jin \textit{et al.}  \cite{jin2020deep} to construct Baseline (Disparity) based on the following facts. Among the compared LF reconstruction methods, Wu \textit{et al.} \cite{wu2019learning} and Wu \textit{et al.} \cite{wu2021revisiting} are both EPI-based methods which do not accept such disparity information. 
Kalantari \textit{et al.} \cite{kalantari2016learning} and Jin \textit{et al.} \cite{jin2020deep} are both warping-based methods which need such information, and Jin \textit{et al.} \cite{jin2020deep} can achieve higher performance than Kalantari \textit{et al.} \cite{kalantari2016learning}. 
}, 
so that both our method and Baseline (Disparity) perceive the same input data to achieve a fair comparison. 
To be specific, we modified Jin \textit{et al.} \cite{jin2020deep} by adding a sub-CNN to estimate the disparity map of the novel view from the input disparity maps, and then blending the estimated target disparity map with the one estimated from the plane plane-sweep volumes (PSVs) with confidence maps.

\begin{figure*}[t]
\begin{center}
  \includegraphics[width=\linewidth]{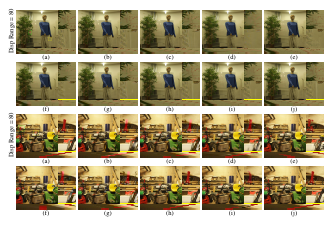}
\end{center}
\vspace{-0.4cm}
  \caption{Visual comparisons of reconstructed SAIs from different methods on the MPI LF dataset \cite{Vamsi2017}. 
  (a) Ground Truth, (b) Baseline (Warp), (c) Baseline (Disparity), (d) Kalantari \textit{et al.} \cite{kalantari2016learning}, (e) Wu \textit{et al.} \cite{wu2019learning}, (f) Wu \textit{et al.} \cite{wu2021revisiting}, (g) Jin \textit{et al.} \cite{jin2020deep}, (h) Guo \textit{et al.} \cite{guo2021learning}, (i) Bao \textit{et al.} \cite{bao2019depth}, (j) Ours. The disparity range between input SAIs reaches $80$ pixels for each reconstructed LF.}
\label{fig:visual_mpi}
\end{figure*}

\subsubsection{Quantitative comparisons on the Inria Sparse LF dataset}
  Table~\ref{table:quantitative} lists the quantitative comparison of different methods on the Inria Sparse dataset, where it can be observed that:
\begin{itemize}
    \item our method improves the average PSNR of the preliminary version, i.e., Guo \textit{et al.} \cite{guo2021learning}, by more than $0.8$ dB, which 
    is credited to the newly proposed components, i.e., the global content information, the adaptive PSV fusion, the feature-space warping, and the weight-smoothness loss term. See Section \ref{ablation_study} for the detailed ablation studies on these components;  
    
    \item our method achieves significantly higher PSNR and SSIM than Baseline (Warp), which directly verifies the the advantage of  the proposed content-aware  warping over the traditional warping operation;

    \item our method achieves higher performance than both Wu \textit{et al.} \cite{wu2019learning} and Wu \textit{et al.} \cite{wu2021revisiting}. The reason may be that 
    they perform reconstruction on 
    2-D EPIs without sufficient modeling of 
    the spatial domain of each view, 
    while our method employs a feature-assistant refinement module to refine the spatial correlation among pixels of novel views; 
    
    \item our method achieves higher PSNR and SSIM than both Kalantari \textit{et al.} \cite{kalantari2016learning} and Jin \textit{et al.} \cite{jin2020deep}. The main reason may be that in addition to the natural limitations of the adopted traditional warping operation, they also cannot provide effective refinement on LFs with large disparities. However, our method   overcomes the limitations by the content-aware warping  and effectively propagates the spatial correlation to the novel view via the feature-assistant refinement module.  Besides, our method achieves higher performance than Baseline (Disparity), demonstrating that the advantage of our framework does not completely come from adopting more accurate disparity maps; and

    \item our method achieves higher performance than Bao \textit{et al.} \cite{bao2019depth} on all scenes except $Flowers$. Although Bao \textit{et al.} \cite{bao2019depth} also proposed a learned warping layer, it has to first explicitly specify the center of the interpolation neighborhood based on the optical flow and depth before conducting the interpolation. The synthesis quality would highly depend on the accuracy of the optical flow and depth estimation. Conversely, our method embeds the estimated optical flow into the MLP as the geometry code, which can tolerate the error of the optical flow estimation to some extent.

\end{itemize}

\begin{figure*}[thp]
\begin{center}
  \includegraphics[width=\linewidth]{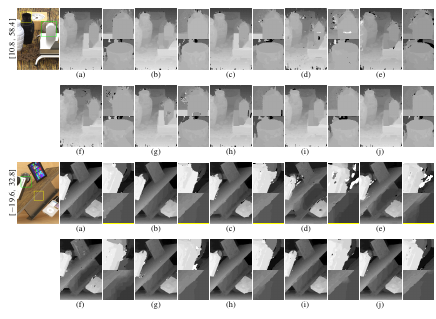}
\end{center}
\vspace{-0.4cm}
  \caption{Visual comparisons of estimated depth maps from ground-truth LFs and reconstructed LF by different methods on the Inria Sparse LF dataset \cite{shi2019framework}. 
  (a) Ground Truth, (b) Baseline (Warp), (c) Baseline (Disparity), (d) Kalantari \textit{et al.} \cite{kalantari2016learning}, (e) Wu \textit{et al.} \cite{wu2019learning}, (f) Wu \textit{et al.} \cite{wu2021revisiting}, (g) Jin \textit{et al.} \cite{jin2020deep}, (h) Guo \textit{et al.} \cite{guo2021learning}, (i) Bao \textit{et al.} \cite{bao2019depth}, (j) Ours. The disparity range between input SAIs of each LF is shown on the left.}
\label{fig:visual_inria_depth}
\end{figure*}

\subsubsection{Quantitative comparisons on the MPI LF dataset}
We also evaluated different methods under different disparity ranges on the MPI dataset \cite{Vamsi2017}. Each scene contains a high-angular densely-sampled LF image composed of $101$ SAIs distributed on a scanline with spatial resolution $720\times 960$. The disparity between adjacent SAI is around $1$ pixel. We  sampled SAIs with different intervals along the angular dimension to construct LFs with different disparity ranges. Specifically, we separately set $10$ disparity ranges from $8$ to $80$ pixels between two input SAIs. For each disparity range, we evenly sampled $3$ SAIs between two input SAIs as ground truth. From Fig.~\ref{fig:quanti_mpi}, it can be seen that the performance of all methods decreases with the disparity range increasing because the reconstruction problem is more challenging,  
but our method  consistently achieves the highest PSNR and SSIM among 
all methods under each disparity range, demonstrating the robustness of our method towards different disparity ranges.

\subsubsection{Comparisons of visual results}
We visually compared the reconstructed LFs by different methods 
in Figs.~\ref{fig:visual_inria} and~\ref{fig:visual_mpi}, where it can be observed that our method can produce views with sharp edges at the occlusion boundaries that are closer to ground truth, while the other  methods produce views with either severe distortions or heavy blurry effects at these regions. Besides, our method can produce better high-frequency details at regions with rich textures than other methods. 
We refer the readers to the \textit{Supplementary Material} for more visual results.

\subsubsection{Comparisons of the LF parallax structure}
As the parallax structure is one of the most important values of LF data, 
we thus managed to compare the parallax structures of LFs reconstructed by different methods. 
First, as shown in Figs.~\ref{fig:visual_inria} and \ref{fig:visual_mpi}, for the formed EPIs, our method can preserve clearer linear structures than other methods, even for lines corresponding to regions with large disparities, demonstrating the strong ability of our method in preserving the parallax structure on extremely sparse LFs. 
Generally, depth maps estimated from higher quality LFs shall be closer to those estimated from ground-truth ones. Thus, we further compared the depth maps estimated from reconstructed LFs by different methods via an identical LF depth estimation method \cite{wang2015occlusion}.  As shown in Fig.~\ref{fig:visual_inria_depth}, our method can produce depth maps with sharper edges at occlusion boundaries and preserve smoothness at regions with uniform depth, which are closest to the ground truth ones. Such observations also demonstrate the advantage of our method on preserving the LF parallax structure. 
Besides, as the ground-truth depth maps of the testing LF images are available, we also quantitatively compared the accuracy of depth estimated from reconstructed LFs by different methods. 
To be specific, we calculated the Mean Square error (MSE) 
between the estimated depth maps and ground-truth ones.  As shown in Table \ref{table:depth}, our method produces lower MSE 
values than all of the compared methods. 

\begin{table*}[t]
\centering
\setlength{\tabcolsep}{2mm}
\caption{Quantitative comparisons 
of the depth maps estimated from the ground-truth LFs and the reconstructed LFs by different methods on the Inria Sparse LF dataset \cite{shi2019framework}. The best and second best results are highlighted in \textcolor{red}{red} and \textcolor{blue}{blue}, respectively.}
\resizebox{\textwidth}{!}{
\begin{tabular}{ccccccccccc}
\toprule
&\makecell{Baseline\\(Warp)} &\makecell{Baseline\\(Disparity)} &\makecell{Kalantari \textit{et al.}\\\cite{kalantari2016learning}}  &\makecell{Wu \textit{et al.}\\\cite{wu2019learning}} &\makecell{Wu \textit{et al.}\\\cite{wu2021revisiting}} &\makecell{Jin \textit{et al.}\\\cite{jin2020deep}} &\makecell{Guo \textit{et al.}\\\cite{guo2021learning}} &\makecell{Bao \textit{et al.}\\\cite{bao2019depth}} &Ours & GT\\ 
\midrule
MSE &19.23 &18.98 &23.39 &24.92 &22.29 &21.28 &18.15 &19.06 &\textcolor{blue}{18.06} &\textcolor{red}{16.96}\\
\bottomrule
\end{tabular}}
\label{table:depth}
\end{table*}

\subsubsection{Efficiency comparison}
We compared the efficiency and model size of different methods.  We implemented all the methods on a Linux server with Intel CPU E5-2699 @ 2.20GHz, 128GB RAM and Tesla V100. As listed in Table~\ref{table:lf_efficiency}, we can see that our method is much faster than Wu \textit{et al.} \cite{wu2019learning} but slower than the other methods. Besides, our model size is smaller than Bao \textit{et al.} \cite{bao2019depth}, but larger than the other methods. 
Although our method is able to reconstruct novel views with much higher quality, there is still room to improve its efficiency.

\begin{table*}[t]
\centering
\setlength{\tabcolsep}{2mm}
\caption{Comparisons of running time (in seconds per view) and model parameter size (M) of different methods on the Inria Sparse LF dataset \cite{shi2019framework}. 
}
\resizebox{\textwidth}{!}{
\begin{tabular}{cccccccccc}
\toprule
&\makecell{Baseline\\(Warp)} &\makecell{Baseline\\(Disparity)} &\makecell{Kalantari \textit{et al.}\\\cite{kalantari2016learning}}  &\makecell{Wu \textit{et al.}\\\cite{wu2019learning}} &\makecell{Wu \textit{et al.}\\\cite{wu2021revisiting}} &\makecell{Jin \textit{et al.}\\\cite{jin2020deep}} &\makecell{Guo \textit{et al.}\\\cite{guo2021learning}} &\makecell{Bao \textit{et al.}\\\cite{bao2019depth}} &Ours \\
\midrule
Time  &7.64 &0.82 &6.32 &27.77 &2.82 &0.69 &2.78 &0.21 &11.01\\
\# Params  &7.25 &2.52 &2.55 & 0.24 &0.55 &2.22 &0.69 &24.03 &7.45 \\
\bottomrule
\end{tabular}}
\label{table:lf_efficiency}
\end{table*}

\subsection{Evaluation on Multi-view Datasets}

\begin{table*}[t]
\centering
\caption{Quantitative comparisons (PSNR/SSIM) of different methods on the DTU dataset \cite{aanaes2016large}. The best and second best results are highlighted in \textcolor{red}{red} and \textcolor{blue}{blue}, respectively.}
\resizebox{0.9\textwidth}{!}{
\begin{tabular}{c|cccccc}
\toprule
Scenes &FVS \cite{riegler2020free} &pixelNeRF \cite{yu2021pixelnerf} & IBRNet \cite{wang2021ibrnet} & SVNVS \cite{shi2021self} & Guo \textit{et al.} \cite{guo2021learning} &Ours 
\\ \midrule
scan3 & 16.52/0.602 & 15.05/0.421 & 18.50/0.606 & \textcolor{blue}{20.01}/\textcolor{red}{0.729} & 18.80/0.621 & \textcolor{red}{20.49}/\textcolor{blue}{0.707} \\ 
scan5 & 16.30/0.575 & 16.30/0.562 & 20.65/0.658 & \textcolor{blue}{20.75}/\textcolor{red}{0.752} & 19.13/0.632 & \textcolor{red}{21.40}/\textcolor{blue}{0.733} \\ 
scan17 & 16.50/0.552 & 14.53/0.450 & 17.81/0.583 & \textcolor{blue}{18.81}/\textcolor{red}{0.686} & 17.27/0.532 & \textcolor{red}{19.57}/\textcolor{blue}{0.669} \\ 
scan21 & 15.28/0.525 & 13.43/0.404 & 15.88/0.520 & \textcolor{red}{16.91}/\textcolor{red}{0.654} & 15.07/0.492 & \textcolor{blue}{16.73}/\textcolor{blue}{0.603} \\ 
scan28 & 17.20/0.603 & 14.77/0.519 & 17.64/0.635 & \textcolor{red}{19.18}/\textcolor{red}{0.744} & 15.10/0.528 & \textcolor{blue}{18.14}/\textcolor{blue}{0.692} \\ 
scan35 & 21.09/0.723 & 13.66/0.571 & \textcolor{blue}{21.26}/0.733 & 19.51/\textcolor{blue}{0.766} & 20.42/0.714 & \textcolor{red}{22.91}/\textcolor{red}{0.787} \\ 
scan37 & 18.57/0.670 & 14.89/0.647 & 19.52/0.713 & \textcolor{blue}{20.56}/\textcolor{blue}{0.787} & 18.62/0.710 & \textcolor{red}{21.22}/\textcolor{red}{0.795} \\ 
scan38 & 18.73/0.591 & 15.92/0.554 & 20.74/0.634 & \textcolor{blue}{21.80}/\textcolor{red}{0.749} & 20.11/0.612 & \textcolor{red}{22.52}/\textcolor{blue}{0.730} \\ 
scan40 & 19.03/0.588 & 16.21/0.596 & 20.97/0.646 & \textcolor{blue}{21.38}/\textcolor{red}{0.749} & 20.59/0.624 & \textcolor{red}{21.41}/\textcolor{blue}{0.720} \\
scan43 & 17.34/0.671 & 16.22/0.568 & 19.83/0.705 & \textcolor{blue}{20.57}/\textcolor{blue}{0.778} & 18.85/0.676 & \textcolor{red}{21.21}/\textcolor{red}{0.786} \\ 
scan56 & 21.35/0.678 & 23.89/0.718 & 24.00/0.682 & \textcolor{blue}{24.08}/\textcolor{red}{0.741} & 22.31/0.611 & \textcolor{red}{24.10}/\textcolor{blue}{0.733} \\ 
scan59 & 18.06/0.699 & 16.90/0.660 & 20.32/0.769 & \textcolor{blue}{21.84}/\textcolor{blue}{0.832} & 17.53/0.685 & \textcolor{red}{22.35}/\textcolor{red}{0.837} \\ 
scan66 & 21.06/0.778 & 21.75/0.806 & \textcolor{red}{25.09}/0.816 & \textcolor{blue}{24.45}/\textcolor{blue}{0.828} & 22.24/0.781 & 23.57/\textcolor{red}{0.833} \\ 
scan67 & 19.45/0.751 & 20.33/0.802 & \textcolor{red}{25.00}/0.807 & \textcolor{blue}{22.83}/\textcolor{blue}{0.808} & 21.47/0.767 & 22.79/\textcolor{red}{0.813} \\ 
scan82 & 19.54/0.788 & 20.44/0.834 & \textcolor{red}{22.83}/\textcolor{blue}{0.854} & 22.31/0.851 & 19.98/0.791 & \textcolor{blue}{22.69}/\textcolor{red}{0.862} \\ 
scan86 & 26.29/0.767 & 26.62/0.797 & 28.58/0.797 & \textcolor{blue}{29.78}/\textcolor{red}{0.822} & 28.35/0.776 & \textcolor{red}{29.95}/\textcolor{blue}{0.816} \\ 
scan106 & 23.43/0.791 & 22.73/0.779 & \textcolor{blue}{25.68}/0.815 & \textcolor{red}{26.58}/\textcolor{red}{0.835} & 24.28/0.774 & 25.52/\textcolor{blue}{0.832} \\
scan117 & 22.73/0.776 & 22.87/0.780 & \textcolor{blue}{26.69}/0.818 & 26.47/\textcolor{blue}{0.829} & 25.16/0.780 & \textcolor{red}{26.71}/\textcolor{red}{0.833} \\ \hline
Average & 19.36/0.674 & 18.14/0.637 & 21.72/0.711 & \textcolor{blue}{22.10}/\textcolor{red}{0.774} & 20.29/0.672 & \textcolor{red}{22.41}/\textcolor{blue}{0.766} \\  
\bottomrule
\end{tabular}}
\label{table:multiview_psnrssim_dtu}
\end{table*}

\begin{table}[thp]
\centering
\setlength{\tabcolsep}{1.5mm}
\caption{Quantitative comparisons (PSNR/SSIM) of different methods on the RealEstate10K dataset \cite{zhou2018stereo}. The best and second best results are highlighted in \textcolor{red}{red} and \textcolor{blue}{blue}, respectively.}
\begin{tabular}{c|cccc}
\toprule
Scenes  &IBRNet \cite{wang2021ibrnet} &SVNVS \cite{shi2021self} &Guo \textit{et al.} \cite{guo2021learning} &Ours \\ 
\midrule
RE10K-1 & 35.58/0.976 & 32.77/0.959 & \textcolor{blue}{39.76}/\textcolor{blue}{0.986} & \textcolor{red}{40.09}/\textcolor{red}{0.987} \\ 
RE10K-2 & 28.04/0.954 & 22.42/0.873 & \textcolor{blue}{31.91}/\textcolor{red}{0.971} & \textcolor{red}{31.94}/\textcolor{blue}{0.970} \\ 
RE10K-3 & 35.44/0.994 & 28.76/0.948 & \textcolor{red}{46.96}/\textcolor{red}{0.996} & \textcolor{blue}{45.31}/\textcolor{blue}{0.995} \\ 
RE10K-4 & 27.11/0.915 & 24.14/0.852 & \textcolor{blue}{29.32}/\textcolor{blue}{0.947} & \textcolor{red}{29.81}/\textcolor{red}{0.955} \\ 
RE10K-5 & 29.71/0.892 & 28.73/0.919 & \textcolor{blue}{35.14}/\textcolor{blue}{0.958} & \textcolor{red}{36.12}/\textcolor{red}{0.966} \\ 
RE10K-6 & 33.56/0.955 & 32.06/0.959 & \textcolor{blue}{38.49}/\textcolor{blue}{0.982} & \textcolor{red}{39.03}/\textcolor{red}{0.984} \\ 
RE10K-7 & 27.61/0.836 & 26.48/0.861 & \textcolor{red}{31.93}/\textcolor{red}{0.930} & \textcolor{blue}{31.11}/\textcolor{blue}{0.915} \\ 
RE10K-8 & 30.19/0.917 & 24.51/0.914 & \textcolor{red}{35.21}/\textcolor{red}{0.963} & \textcolor{blue}{34.66}/\textcolor{blue}{0.960} \\ 
RE10K-9 & 28.65/0.929 & 28.98/0.920 & \textcolor{blue}{31.74}/\textcolor{blue}{0.957} & \textcolor{red}{32.00}/\textcolor{red}{0.961} \\ 
RE10K-10 & 23.28/0.772 & 22.99/0.820 & \textcolor{red}{27.55}/\textcolor{blue}{0.895} & \textcolor{blue}{27.53}/\textcolor{red}{0.898} \\ 
RE10K-11 & 24.26/0.882 & 22.88/0.845 & \textcolor{blue}{25.45}/\textcolor{blue}{0.888} & \textcolor{red}{25.53}/\textcolor{red}{0.891} \\ 
RE10K-12 & 27.06/0.916 & 30.66/0.955 & \textcolor{blue}{36.10}/\textcolor{blue}{0.982} & \textcolor{red}{36.94}/\textcolor{red}{0.986} \\ 
RE10K-13 & 31.47/0.947 & 32.34/0.960 & \textcolor{blue}{38.48}/\textcolor{blue}{0.984} & \textcolor{red}{38.50}/\textcolor{red}{0.984} \\ 
RE10K-14 & 25.05/0.851 & 29.52/0.943 & \textcolor{blue}{33.76}/\textcolor{blue}{0.972} & \textcolor{red}{35.10}/\textcolor{red}{0.979} \\ 
RE10K-15 & 26.87/0.862 & 24.55/0.886 & \textcolor{blue}{29.88}/\textcolor{blue}{0.939} & \textcolor{red}{30.82}/\textcolor{red}{0.949} \\ 
RE10K-16 & 35.10/0.971 & 34.36/0.957 & \textcolor{blue}{41.48}/\textcolor{blue}{0.984} & \textcolor{red}{42.03}/\textcolor{red}{0.986} \\
RE10K-17 & 35.87/0.975 & 31.34/0.960 & \textcolor{blue}{38.63}/\textcolor{blue}{0.984} & \textcolor{red}{39.97}/\textcolor{red}{0.987} \\ \hline
Average & 29.70/0.914 & 28.09/0.914 & \textcolor{blue}{34.81}/\textcolor{blue}{0.960} & \textcolor{red}{35.09}/\textcolor{red}{0.962} \\  
\bottomrule
\end{tabular}
\label{table:multiview_psnrssim_reastate}
\end{table}

We evaluated our method on two  multi-view datasets, i.e., DTU  \cite{aanaes2016large} and RealEstate10K \cite{zhou2018stereo}. The transformation between views in these datasets has more degrees of freedom than that of LF datasets, leading to more complicated parallax structure, and thus more challenging for view synthesis methods. On the DTU dataset \cite{aanaes2016large},  we compared the proposed method with three IBR methods named FVS \cite{riegler2020free}, SVNVS \cite{shi2021self}, and Guo \textit{et al.} \cite{guo2021learning}, as well as two NeRF-based methods named pixelNeRF \cite{yu2021pixelnerf} and IBRNet \cite{wang2021ibrnet}. On the RealEstate10K dataset \cite{zhou2018stereo}, we compared the proposed method with SVNVS \cite{shi2021self}, Guo \textit{et al.} \cite{guo2021learning}, and IBRNet \cite{wang2021ibrnet}. We did not compare the proposed method with FVS \cite{riegler2020free} and pixelNeRF \cite{yu2021pixelnerf} on the RealEstate10K  dataset \cite{zhou2018stereo}  since they require either off-the-shelf depth maps or scale matrix which are not provided by the RealEstate10K dataset \cite{zhou2018stereo}. For fair comparisons, we trained the compared methods on the same dataset as ours with the officially released codes and suggested configurations.

\subsubsection{Quantitative comparison}
We quantitatively compared different methods in terms of PSNR and SSIM in Tables \ref{table:multiview_psnrssim_dtu} and   \ref{table:multiview_psnrssim_reastate}, where it can be observed that
\begin{itemize}
    \item our method achieves higher average PSNR and SSIM values than our preliminary conference version Guo \textit{et al.} \cite{guo2021learning} on both DTU \cite{aanaes2016large} and RealEstate10K \cite{zhou2018stereo}, and especially the improvement on the DTU dataset with more complicated parallax structures is more than 2 dB, which is credited to the newly proposed global content information, the adaptive PSV fusion, the feature-space warping, and the weight smoothness loss term. 

    \item  our method achieves higher PSNR and SSIM values than FVS \cite{riegler2020free} and SVNVS \cite{shi2021self}. The reason is  that both FVS and SVNVS employ the traditional warping operation in their IBR frameworks, which leads to distortions and blurry effects at texture edges and occlusion boundaries.
    On the contrary, our method employs content-aware warping that can produce high-quality results; and 
    
    \item our method achieves much higher PSNR and SSIM values than pixelNeRF \cite{yu2021pixelnerf} and IBRNet \cite{wang2021ibrnet}. For each query point in the neural field, these methods employ simple bilinear interpolation to extract colors and features at the projected pixel locations in source views, and thus, the accuracy of estimated density and color would be limited.
    By contrast, our method can adaptively locate corresponding pixels in source views by learning interpolation weights.
\end{itemize}

\begin{figure*}[thp]
\begin{center}
  \includegraphics[width=0.9\linewidth]{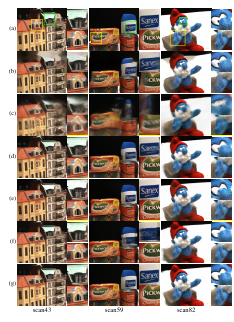}
\end{center}
\vspace{-0.4cm}
\caption{ Visual comparisons of synthesized views from different methods on the DTU dataset \cite{aanaes2016large}. (a) Ground Truth, (b) FVS \cite{riegler2020free}, (c) pixelNeRF \cite{yu2021pixelnerf}, (d) IBRNet \cite{wang2021ibrnet}, (e) SVNVS \cite{shi2021self}, (f) Guo \textit{et al.} \cite{guo2021learning}, and (g) Ours.}
\label{fig:visual_dtu}
\end{figure*}

\begin{figure*}[thp]
\begin{center}
  \includegraphics[width=\linewidth]{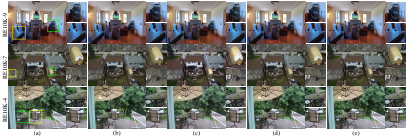}
\end{center}
\vspace{-0.4cm}
\caption{Visual comparisons of synthesized views from different methods on the RealEstate10K dataset \cite{zhou2018stereo}. (a) Ground Truth, (b) IBRNet \cite{wang2021ibrnet}, (c) SVNVS \cite{shi2021self}, (d) Guo \textit{et al.} \cite{guo2021learning}  and (e) Ours.}
\label{fig:visual_realestate}
\end{figure*}

\subsubsection{Comparisons of visual results}
We compared the visual results of different methods on both the DTU dataset \cite{aanaes2016large} and the RealEstate10K dataset \cite{zhou2018stereo} in Figs.~\ref{fig:visual_dtu} and~\ref{fig:visual_realestate}, which further demonstrate the advantage of our proposed method. As can be seen, our method can synthesize high-quality details and structures at most areas, while the compared methods either produce severe distortions or blurry effects at these regions. Besides, 
our method can preserve sharp edges at occlusion boundaries, while other methods show severe ghost effects.
We refer the readers to the \textit{Supplementary Material} for more visual results.

\begin{table}[t]
\centering
\setlength{\tabcolsep}{1.3mm}
\caption{Comparisons of running time (in seconds per view) and model parameter size (M) of different methods on the DTU dataset \cite{aanaes2016large}. 
}
\begin{tabular}{ccccccc}
\toprule
&\makecell{FVS \\\cite{riegler2020free}}  &\makecell{pixelNeRF \\\cite{yu2021pixelnerf}}  &\makecell{IBRNet \\\cite{wang2021ibrnet}} &\makecell{SVNVS \\\cite{shi2021self}} &\makecell{Guo \textit{et al.} \\\cite{guo2021learning}} &Ours \\ 
\midrule
Time  & 0.14 &14.29 &2.82 &0.15 &1.02 &6.72 \\
\# Params  &33.73 &28.16 &8.96 &153.15 &0.69 &7.47 \\
\bottomrule
\end{tabular}
\label{table:multiview_efficiency}
\end{table}

\subsubsection{Comparisons of efficiency}
We compared the efficiency and model size of different methods.  We implemented all methods on a Linux server with Intel CPU E5-2620 @ 2.10GHz, 256GB RAM and GeForce RTX 2080 Ti. As listed in Table~\ref{table:multiview_efficiency}, our method 
is faster than pixelNeRF \cite{yu2021pixelnerf} while slower than the other compared methods, and our model size is larger than Guo \textit{et al.} \cite{guo2021learning} while smaller than other methods. Taking the reconstruction quality, efficiency, and model size together, we believe our method is the best.

\subsubsection{Comparisons of different numbers of input views}

\begin{table}[thp]
\centering
\setlength{\tabcolsep}{3mm}
\caption{Quantitative performance (PSNR/SSIM) of our method fed with various numbers of source views on the DTU dataset dataset \cite{aanaes2016large}.}
\begin{tabular}{c|ccc}
\toprule
Scenes & 2-inputs & 3-inputs & 4-inputs \\ \midrule
scan3 & 20.49/0.707 & 20.76/0.747 & 21.07/0.764 \\ 
scan5 & 21.40/0.733 & 21.79/0.776 & 21.96/0.790 \\ 
scan17 & 19.57/0.669 & 20.42/0.726 & 21.16/0.752 \\ 
scan21 & 16.73/0.603 & 18.69/0.696 & 19.43/0.730 \\ 
scan28 & 18.14/0.692 & 19.73/0.748 & 20.38/0.773 \\ 
scan35 & 22.91/0.787 & 23.47/0.808 & 23.76/0.818 \\ 
scan37 & 21.22/0.795 & 21.76/0.820 & 22.24/0.832 \\ 
scan38 & 22.52/0.730 & 23.48/0.770 & 23.93/0.787 \\ 
scan40 & 21.41/0.720 & 22.68/0.764 & 23.16/0.782 \\ 
scan43 & 21.21/0.786 & 22.07/0.824 & 22.67/0.844 \\ 
scan56 & 24.10/0.733 & 25.20/0.791 & 25.77/0.810 \\ 
scan59 & 22.35/0.837 & 23.13/0.857 & 23.41/0.871 \\ 
scan66 & 23.57/0.833 & 24.38/0.860 & 24.52/0.868 \\ 
scan67 & 22.79/0.813 & 23.48/0.836 & 23.73/0.841 \\ 
scan82 & 22.69/0.862 & 23.28/0.880 & 23.76/0.890 \\ 
scan86 & 29.95/0.816 & 30.38/0.825 & 30.48/0.830 \\ 
scan106 & 25.52/0.832 & 26.39/0.862 & 27.18/0.873 \\ 
scan117 & 26.71/0.833 & 27.70/0.866 & 28.27/0.877 \\ \hline
\textbf{Average} & 22.41/0.766 & 23.27/0.803 & 23.72/0.819 \\ 
\bottomrule
\end{tabular}
\label{table:input_number}
\end{table}

We evaluated the proposed method 
with 2, 3 and 4 source views fed on the DTU dataset \cite{aanaes2016large}. 
Specifically, in addition to selecting two adjacent views of the target view as source views, we separately added one and two views as the additional source views for the 3-inputs and 4-inputs tasks, respectively, by using the view selection result provided by \cite{yao2018mvsnet}.
As listed in Table \ref{table:input_number}, it can be observed that our method can achieve better reconstruction quality along with the number of input views increasing. More specifically, the improvement is more obvious from 2-inputs to 3-inputs than that of from 3-inputs to 4-inputs. The reason may be that 3 inputs provide much more useful information than 2 inputs and is sufficient for the reconstructions of most target views. Adding another source view based on 3 inputs, i.e., 4 inputs, could not provide much useful information.


\begin{figure*}[thp]
\begin{center}
  \includegraphics[width=\linewidth]{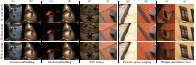}
\end{center}
\caption{Visual illustration of the ablation studies on the five key components of our framework.}
\label{fig:visual_ablation}
\end{figure*}

\begin{table*}[!t]
\centering
\setlength{\tabcolsep}{1.5mm}
\caption{Quantitative results of the ablation studies on the five key components of our method. ``$\times$" denotes that the corresponding component is not included, while ``$\surd$"  denotes being included. 
}
\resizebox{0.8\textwidth}{!}{
\begin{tabular}{c|ccccc|c}
\toprule
Model number&\makecell{Content\\embedding} & \makecell{Global\\embedding} & \makecell{PSV fusion}  & \makecell{Feature-space\\ warping} & \makecell{Weight-smoothness\\loss} &PSNR/SSIM\\
\midrule
1& $\times$ & $\times$ & $\times$ & $\times$ & $\times$ & 20.69/0.704 \\ 
2& $\surd$ & $\times$ & $\times$ & $\times$ & $\times$ & 21.85/0.750 \\ 
3&$\surd$ & $\surd$ & $\times$ & $\times$ & $\times$ & 21.94/0.751 \\ 
4&$\surd$ & $\surd$ & $\surd$ & $\times$ & $\times$ & 22.15/0.759 \\ 
5&$\surd$ & $\surd$ & $\surd$ & $\surd$ & $\times$ & 22.30/0.767 \\ 
6&$\surd$ & $\surd$ & $\surd$ & $\surd$ & $\surd$ & 22.41/0.766 \\ 
\bottomrule
\end{tabular}}
\label{table:ablation}
\end{table*}

\subsection{Ablation Study}
\label{ablation_study}

To validate the effectiveness of the key components of our method, i.e., content embedding, global embedding, adaptive PSV fusion, feature-space
warping, and weight-smoothness loss, we carried out comprehensive ablation studies on the DTU
dataset \cite{aanaes2016large} under the setting of 2 input views. 
To be specific, we sequentially added the modules to the base model one by one until all the five components were included to form the complete model. 

\textbf{Content embedding.} By comparing the results of Models \#1 and \#2 in Table~\ref{table:ablation}, it can be seen that there is a significant increase of performance when adding the content embedding to the base model, which verifies the advantage brought by detecting the texture edges of input views, and understanding the occlusion and non-Lambertian relations between input views. Figs.~\ref{fig:visual_ablation} (a) and (b) also  visually verify 
 the advantage.

\textbf{Global embedding.}
By comparing the results of Models \#2 and \#3 in Table~\ref{table:ablation}, we can see that embedding the global information in the content-aware warping can improve the synthesis performance. As shown in  Figs.~\ref{fig:visual_ablation} (c) and (d), we can see that the results of the framework with global embedding are closer to the ground truth at the occlusion boundary and reflecting area.

\textbf{PSV fusion.} According to the results of  Models \#3 and \#4 in Table~\ref{table:ablation}, this component involved in the spatial refinement module can improve the PSNR value by about $0.19$ dB.  As shown in Figs.~\ref{fig:visual_ablation} (e) and (f), the results without this component are obviously blurry at the occlusion boundaries and the texture regions. 

\textbf{Feature-space warping.}
We can validate the advantage of this component contained in the spatial refinement module by comparing the results of Models \#4 and \#5 in Table~\ref{table:ablation}. As shown in Figs.~\ref{fig:visual_ablation} (g) and (h), without this component, some fine structures, such as delicate objects and textures, are obviously broken.

\textbf{Weight-smoothness loss.} The effectiveness of this loss term is validated by comparing the results of Models \#5 and \#6 in Table~\ref{table:ablation}. 
Besides, as shown in Figs.~\ref{fig:visual_ablation} (i) and (j), adding this component generates more visually-pleasing results than being without it, which also verifies the advantage. 

\begin{figure}[thp]
\begin{center}
  \includegraphics[width=\linewidth]{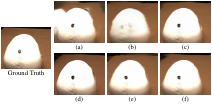}
\end{center}
\vspace{-0.4cm}
  \caption{Visual comparisons of the synthesized view from different methods on the scene with large texture-less regions. (a) FVS \cite{riegler2020free}, (b) pixelNeRF \cite{yu2021pixelnerf}, (c) IBRNet \cite{wang2021ibrnet}, (d) SVNVS \cite{shi2021self}, (e) Guo \textit{et al.} \cite{guo2021learning} and (f) Ours.}
\label{fig:limit_dtu}
\end{figure}

\begin{figure*}[thp]
\begin{center}
  \includegraphics[width=\linewidth]{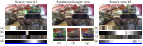}
\end{center}
\vspace{-0.4cm}
  \caption{Neighborhoods and learned weights for synthesizing a typical pixel on the Tanks and Temples dataset \cite{knapitsch2017tanks}. The synthesized pixel is highlighted with a red dot in the synthesized target view. The pixels corresponding to the maximum weight values in the neighborhoods are highlighted with green and yellow dots in source view \#1 and \#2, respectively. Below source views, from top to bottom: 
  (a) and (h) are the zoom-in of the neighbors highlighted with green and yellow straight lines, respectively;  
  (b) and (i) are the learned weights corresponding to (a) and (h), respectively; 
  (c) and (j) are the zoom-in of the neighbors highlighted with blue straight lines; 
  and (d) and (k) are learned weights corresponding to (c) and (j), respectively.
  The regions with red, green, and yellow frames around the synthesized pixel and maximum-weight pixels are also zoomed in, i.e., (e), (f), and (g), for better visualization.}
\label{fig:limit_tat}
\end{figure*}

\subsection{Limitation}

According to Table \ref{table:multiview_psnrssim_dtu}, our method is inferior to IBRNet \cite{wang2021ibrnet} and SVNVS \cite{shi2021self} on some scenes of the DTU dataset, e.g., scan66, scan67 and scan106 which contain many \textit{texture-less} or \textit{repeated texture} regions. For example, as shown in Fig. \ref{fig:limit_dtu}, it can be observed that our method achieves better visual results than FVS \cite{riegler2020free} and pixelNeRF \cite{yu2021pixelnerf}, but worse than IBRNet \cite{wang2021ibrnet} and SVNVS \cite{shi2021self}. The possible reason is that our content-aware warping performs the interpolation only in a local neighborhood and predicts interpolation weights for each neighboring pixel only conditioned on its local information, and thus, the predicted weights are evenly distributed to all the neighboring pixels with very close intensities at these regions, which results in blurring effects.

Besides, the quality of synthesized views by our method on  the scenes with too various relative camera pose patterns between the source view and the target view is still limited. To illustrate this limitation, we retrained and tested our network on the Tanks and Temples dataset \cite{knapitsch2017tanks} under the setting of 2 input views. Note that unlike the DTU \cite{aanaes2016large} and RealEstate10K \cite{zhou2018stereo} datasets, where the distributions of cameras poses are same on all scenes, or the camera is always moving forward, the Tanks and Temples dataset \cite{knapitsch2017tanks} has more complex and irregular camera trajectories, resulting in more various relative camera pose patterns between the source view and the
target view, making it more challenging to synthesize high-quality views. Specifically, following the method FVS \cite{riegler2020free}, we split the Tanks and Temples dataset \cite{knapitsch2017tanks} into training and testing datasets, i.e., 17 out of 21 scenes for training and the remaining 4 scenes for testing. In the training phase, we adopted the selection strategy provided by FVS \cite{riegler2020free}, which counts the number of pixels from the target view that are mapped to the valid source image domain and selects the top two images that are maximize that score as the source images, to randomly select a view from a scene as the target view and two views as the source views. As shown in the Fig. \ref{fig:limit_tat}, it can be observed that our method produces blurry synthesized views. To deeply analyze this issue, we visualized the learned weights by the proposed content-aware warping for synthesizing a typical pixel of the target view, where we can observe that our content-aware warping cannot assign large weight values to the neighbors which are the correspondences of the target pixel or semantically close to the target pixel, and spreads weights evenly over the neighborhood, resulting in blurring effects in the synthesized target view. The possible reason for this limited ability of our method is that the MLP in the content-aware warping
has difficulties to model such various relative camera pose patterns between the source view and the target view through embedding the correspondence relation and content information in our content-aware warping, and thus, the MLP cannot assign proper weight values to neighbors.

\section{Conclusion}
\label{sec:conclusion}

We have presented an end-to-end learning-based framework for novel view synthesis from a set of input source views. Particularly,  we proposed learnable  content-aware warping to overcome the natural limitations of the  traditional image warping operation. 
Owing to the content-aware warping, as well as the  blending and refinement modules that are elaborately designed to handle occlusions and recover spatial correlations, the proposed view synthesis framework reconstructs novel views with much higher quality on both LF datasets and multi-view datasets, compared with state-of-the-art methods.

\ifCLASSOPTIONcaptionsoff
  \newpage
\fi



\normalem
\bibliographystyle{IEEEtran}
\bibliography{egbib.bib}

\begin{thebibliography}{10}
\providecommand{\url}[1]{#1}
\csname url@samestyle\endcsname
\providecommand{\newblock}{\relax}
\providecommand{\bibinfo}[2]{#2}
\providecommand{\BIBentrySTDinterwordspacing}{\spaceskip=0pt\relax}
\providecommand{\BIBentryALTinterwordstretchfactor}{4}
\providecommand{\BIBentryALTinterwordspacing}{\spaceskip=\fontdimen2\font plus
\BIBentryALTinterwordstretchfactor\fontdimen3\font minus
  \fontdimen4\font\relax}
\providecommand{\BIBforeignlanguage}[2]{{%
\expandafter\ifx\csname l@#1\endcsname\relax
\typeout{** WARNING: IEEEtran.bst: No hyphenation pattern has been}%
\typeout{** loaded for the language `#1'. Using the pattern for}%
\typeout{** the default language instead.}%
\else
\language=\csname l@#1\endcsname
\fi
#2}}
\providecommand{\BIBdecl}{\relax}
\BIBdecl

\bibitem{szeliski2010computer}
R.~Szeliski, \emph{Computer vision: algorithms and applications}.\hskip 1em
  plus 0.5em minus 0.4em\relax Springer Science \& Business Media, 2010.

\bibitem{furukawa2009accurate}
Y.~Furukawa and J.~Ponce, ``Accurate, dense, and robust multiview stereopsis,''
  \emph{IEEE Transactions on Pattern Analysis and Machine Intelligence},
  vol.~32, no.~8, pp. 1362--1376, 2009.

\bibitem{schonberger2016structure}
J.~L. Schonberger and J.-M. Frahm, ``Structure-from-motion revisited,'' in
  \emph{IEEE Conference on Computer Vision and Pattern Recognition (CVPR)},
  2016, pp. 4104--4113.

\bibitem{yao2018mvsnet}
Y.~Yao, Z.~Luo, S.~Li, T.~Fang, and L.~Quan, ``Mvsnet: Depth inference for
  unstructured multi-view stereo,'' in \emph{European Conference on Computer
  Vision (ECCV)}, 2018, pp. 767--783.

\bibitem{guo2020accurate}
C.~Guo, J.~Jin, J.~Hou, and J.~Chen, ``Accurate light field depth estimation
  via an occlusion-aware network,'' in \emph{IEEE International Conference on
  Multimedia and Expo (ICME)}, 2020, pp. 1--6.

\bibitem{lfapp2015vr}
F.-C. Huang, K.~Chen, and G.~Wetzstein, ``The light field stereoscope:
  immersive computer graphics via factored near-eye light field displays with
  focus cues,'' \emph{ACM Transactions on Graphics}, vol.~34, no.~4, p.~60,
  2015.

\bibitem{lfapp2017vryu}
J.~Yu, ``A light-field journey to virtual reality,'' \emph{IEEE MultiMedia},
  vol.~24, no.~2, pp. 104--112, 2017.

\bibitem{wei2019vr}
S.-E. Wei, J.~Saragih, T.~Simon, A.~W. Harley, S.~Lombardi, M.~Perdoch,
  A.~Hypes, D.~Wang, H.~Badino, and Y.~Sheikh, ``Vr facial animation via
  multiview image translation,'' \emph{ACM Transactions on Graphics}, vol.~38,
  no.~4, pp. 1--16, 2019.

\bibitem{debevec1996modeling}
P.~E. Debevec, C.~J. Taylor, and J.~Malik, ``Modeling and rendering
  architecture from photographs: A hybrid geometry-and image-based approach,''
  in \emph{Computer Graphics and Interactive Techniques}, 1996, pp. 11--20.

\bibitem{levoy1996light}
M.~Levoy and P.~Hanrahan, ``Light field rendering,'' in \emph{Proceedings of
  the 23rd annual conference on Computer graphics and interactive techniques},
  1996, pp. 31--42.

\bibitem{zhou2018stereo}
T.~Zhou, R.~Tucker, J.~Flynn, G.~Fyffe, and N.~Snavely, ``Stereo magnification:
  learning view synthesis using multiplane images,'' \emph{ACM Transactions on
  Graphics}, vol.~37, no.~4, pp. 1--12, 2018.

\bibitem{mildenhall2019local}
B.~Mildenhall, P.~P. Srinivasan, R.~Ortiz-Cayon, N.~K. Kalantari,
  R.~Ramamoorthi, R.~Ng, and A.~Kar, ``Local light field fusion: Practical view
  synthesis with prescriptive sampling guidelines,'' \emph{ACM Transactions on
  Graphics}, vol.~38, no.~4, pp. 1--14, 2019.

\bibitem{wang2021ibrnet}
Q.~Wang, Z.~Wang, K.~Genova, P.~P. Srinivasan, H.~Zhou, J.~T. Barron,
  R.~Martin-Brualla, N.~Snavely, and T.~Funkhouser, ``Ibrnet: Learning
  multi-view image-based rendering,'' in \emph{IEEE Conference on Computer
  Vision and Pattern Recognition (CVPR)}, 2021, pp. 4690--4699.

\bibitem{chaurasia2013depth}
G.~Chaurasia, S.~Duchene, O.~Sorkine-Hornung, and G.~Drettakis, ``Depth
  synthesis and local warps for plausible image-based navigation,'' \emph{ACM
  Transactions on Graphics}, vol.~32, no.~3, pp. 1--12, 2013.

\bibitem{hedman2016scalable}
P.~Hedman, T.~Ritschel, G.~Drettakis, and G.~Brostow, ``Scalable inside-out
  image-based rendering,'' \emph{ACM Transactions on Graphics}, vol.~35, no.~6,
  pp. 1--11, 2016.

\bibitem{riegler2020free}
G.~Riegler and V.~Koltun, ``Free view synthesis,'' in \emph{European Conference
  on Computer Vision (ECCV)}.\hskip 1em plus 0.5em minus 0.4em\relax Springer,
  2020, pp. 623--640.

\bibitem{shi2021self}
Y.~Shi, H.~Li, and X.~Yu, ``Self-supervised visibility learning for novel view
  synthesis,'' in \emph{IEEE Conference on Computer Vision and Pattern
  Recognition (CVPR)}, 2021, pp. 9675--9684.

\bibitem{guo2021learning}
M.~Guo, J.~Jin, H.~Liu, and J.~Hou, ``Learning dynamic interpolation for
  extremely sparse light fields with wide baselines,'' in \emph{IEEE
  International Conference on Computer Vision (ICCV)}, 2021, pp. 2450--2459.

\bibitem{levin2008understanding}
A.~Levin, W.~T. Freeman, and F.~Durand, ``Understanding camera trade-offs
  through a bayesian analysis of light field projections,'' in \emph{European
  Conference on Computer Vision (ECCV)}.\hskip 1em plus 0.5em minus 0.4em\relax
  Springer, 2008, pp. 88--101.

\bibitem{levin2010linear}
A.~Levin and F.~Durand, ``Linear view synthesis using a dimensionality gap
  light field prior,'' in \emph{IEEE Computer Society Conference on Computer
  Vision and Pattern Recognition (CVPR)}.\hskip 1em plus 0.5em minus
  0.4em\relax IEEE, 2010, pp. 1831--1838.

\bibitem{mitra2012light}
K.~Mitra and A.~Veeraraghavan, ``Light field denoising, light field
  superresolution and stereo camera based refocussing using a gmm light field
  patch prior,'' in \emph{IEEE Computer Society Conference on Computer Vision
  and Pattern Recognition Workshops (CVPRW)}.\hskip 1em plus 0.5em minus
  0.4em\relax IEEE, 2012, pp. 22--28.

\bibitem{marwah2013compressive}
K.~Marwah, G.~Wetzstein, Y.~Bando, and R.~Raskar, ``Compressive light field
  photography using overcomplete dictionaries and optimized projections,''
  \emph{ACM Transactions on Graphics}, vol.~32, no.~4, pp. 1--12, 2013.

\bibitem{shi2014light}
L.~Shi, H.~Hassanieh, A.~Davis, D.~Katabi, and F.~Durand, ``Light field
  reconstruction using sparsity in the continuous fourier domain,'' \emph{ACM
  Transactions on Graphics}, vol.~34, no.~1, pp. 1--13, 2014.

\bibitem{vagharshakyan2017light}
S.~Vagharshakyan, R.~Bregovic, and A.~Gotchev, ``Light field reconstruction
  using shearlet transform,'' \emph{IEEE Transactions on Pattern Analysis and
  Machine Intelligence}, vol.~40, no.~1, pp. 133--147, 2017.

\bibitem{kamal2016tensor}
M.~H. Kamal, B.~Heshmat, R.~Raskar, P.~Vandergheynst, and G.~Wetzstein,
  ``Tensor low-rank and sparse light field photography,'' \emph{Computer Vision
  and Image Understanding}, vol. 145, pp. 172--181, 2016.

\bibitem{wanner2013variational}
S.~Wanner and B.~Goldluecke, ``Variational light field analysis for disparity
  estimation and super-resolution,'' \emph{IEEE Transactions on Pattern
  Analysis and Machine Intelligence}, vol.~36, no.~3, pp. 606--619, 2013.

\bibitem{zhang2015light}
Z.~Zhang, Y.~Liu, and Q.~Dai, ``Light field from micro-baseline image pair,''
  in \emph{IEEE Conference on Computer Vision and Pattern Recognition
  (CVPR)}.\hskip 1em plus 0.5em minus 0.4em\relax IEEE, 2015, pp. 3800--3809.

\bibitem{yoon2015learning}
Y.~Yoon, H.-G. Jeon, D.~Yoo, J.-Y. Lee, and I.~So~Kweon, ``Learning a deep
  convolutional network for light-field image super-resolution,'' in \emph{IEEE
  International Conference on Computer Vision Workshops (ICCVW)}.\hskip 1em
  plus 0.5em minus 0.4em\relax IEEE, 2015, pp. 24--32.

\bibitem{wu2017light}
G.~Wu, M.~Zhao, L.~Wang, Q.~Dai, T.~Chai, and Y.~Liu, ``Light field
  reconstruction using deep convolutional network on epi,'' in \emph{IEEE
  Conference on Computer Vision and Pattern Recognition (CVPR)}.\hskip 1em plus
  0.5em minus 0.4em\relax IEEE, 2017, pp. 6319--6327.

\bibitem{wang2018end}
Y.~Wang, F.~Liu, Z.~Wang, G.~Hou, Z.~Sun, and T.~Tan, ``End-to-end view
  synthesis for light field imaging with pseudo 4dcnn,'' in \emph{European
  Conference on Computer Vision (ECCV)}, 2018, pp. 333--348.

\bibitem{yeung2018fast}
H.~W.~F. Yeung, J.~Hou, J.~Chen, Y.~Y. Chung, and X.~Chen, ``Fast light field
  reconstruction with deep coarse-to-fine modeling of spatial-angular clues,''
  in \emph{European Conference on Computer Vision (ECCV)}, 2018, pp. 137--152.

\bibitem{meng2019high}
N.~Meng, H.~K.-H. So, X.~Sun, and E.~Lam, ``High-dimensional dense residual
  convolutional neural network for light field reconstruction,'' \emph{IEEE
  Transactions on Pattern Analysis and Machine Intelligence}, 2019.

\bibitem{kalantari2016learning}
N.~K. Kalantari, T.-C. Wang, and R.~Ramamoorthi, ``Learning-based view
  synthesis for light field cameras,'' \emph{ACM Transactions on Graphics},
  vol.~35, no.~6, pp. 1--10, 2016.

\bibitem{wu2019learning}
G.~Wu, Y.~Liu, Q.~Dai, and T.~Chai, ``Learning sheared epi structure for light
  field reconstruction,'' \emph{IEEE Transactions on Image Processing},
  vol.~28, no.~7, pp. 3261--3273, 2019.

\bibitem{srinivasan2017learning}
P.~P. Srinivasan, T.~Wang, A.~Sreelal, R.~Ramamoorthi, and R.~Ng, ``Learning to
  synthesize a 4d rgbd light field from a single image,'' in \emph{IEEE
  International Conference on Computer Vision (ICCV)}, 2017, pp. 2243--2251.

\bibitem{jin2020deep}
J.~Jin, J.~Hou, J.~Chen, H.~Zeng, S.~Kwong, and J.~Yu, ``Deep coarse-to-fine
  dense light field reconstruction with flexible sampling and geometry-aware
  fusion,'' \emph{IEEE Transactions on Pattern Analysis and Machine
  Intelligence}, 2020.

\bibitem{penner2017soft}
E.~Penner and L.~Zhang, ``Soft 3d reconstruction for view synthesis,''
  \emph{ACM Transactions on Graphics}, vol.~36, no.~6, pp. 1--11, 2017.

\bibitem{hedman2018deep}
P.~Hedman, J.~Philip, T.~Price, J.-M. Frahm, G.~Drettakis, and G.~Brostow,
  ``Deep blending for free-viewpoint image-based rendering,'' \emph{ACM
  Transactions on Graphics}, vol.~37, no.~6, pp. 1--15, 2018.

\bibitem{choi2019extreme}
I.~Choi, O.~Gallo, A.~Troccoli, M.~H. Kim, and J.~Kautz, ``Extreme view
  synthesis,'' in \emph{IEEE International Conference on Computer Vision
  (ICCV)}, 2019, pp. 7781--7790.

\bibitem{flynn2019deepview}
J.~Flynn, M.~Broxton, P.~Debevec, M.~DuVall, G.~Fyffe, R.~Overbeck, N.~Snavely,
  and R.~Tucker, ``Deepview: View synthesis with learned gradient descent,'' in
  \emph{IEEEF Conference on Computer Vision and Pattern Recognition (CVPR)},
  2019, pp. 2367--2376.

\bibitem{srinivasan2019pushing}
P.~P. Srinivasan, R.~Tucker, J.~T. Barron, R.~Ramamoorthi, R.~Ng, and
  N.~Snavely, ``Pushing the boundaries of view extrapolation with multiplane
  images,'' in \emph{IEEE Conference on Computer Vision and Pattern Recognition
  (CVPR)}, 2019, pp. 175--184.

\bibitem{tucker2020single}
R.~Tucker and N.~Snavely, ``Single-view view synthesis with multiplane
  images,'' in \emph{IEEE Conference on Computer Vision and Pattern Recognition
  (CVPR)}, 2020, pp. 551--560.

\bibitem{mildenhall2020nerf}
B.~Mildenhall, P.~P. Srinivasan, M.~Tancik, J.~T. Barron, R.~Ramamoorthi, and
  R.~Ng, ``Nerf: Representing scenes as neural radiance fields for view
  synthesis,'' in \emph{European Conference on Computer Vision (ECCV)}.\hskip
  1em plus 0.5em minus 0.4em\relax Springer, 2020, pp. 405--421.

\bibitem{yu2021pixelnerf}
A.~Yu, V.~Ye, M.~Tancik, and A.~Kanazawa, ``pixelnerf: Neural radiance fields
  from one or few images,'' in \emph{IEEE Conference on Computer Vision and
  Pattern Recognition (CVPR)}, 2021, pp. 4578--4587.

\bibitem{porter1984compositing}
T.~Porter and T.~Duff, ``Compositing digital images,'' in \emph{Computer
  Graphics and Interactive Techniques}, 1984, pp. 253--259.

\bibitem{kajiya1984ray}
J.~T. Kajiya and B.~P. Von~Herzen, ``Ray tracing volume densities,'' \emph{ACM
  SIGGRAPH computer graphics}, vol.~18, no.~3, pp. 165--174, 1984.

\bibitem{jaderberg2015spatial}
M.~Jaderberg, K.~Simonyan, A.~Zisserman \emph{et~al.}, ``Spatial transformer
  networks,'' \emph{Advances in neural information processing systems
  (NeurIPS)}, vol.~28, pp. 2017--2025, 2015.

\bibitem{bako2017kernel}
S.~Bako, T.~Vogels, B.~McWilliams, M.~Meyer, J.~Nov{\'a}k, A.~Harvill, P.~Sen,
  T.~Derose, and F.~Rousselle, ``Kernel-predicting convolutional networks for
  denoising monte carlo renderings.'' \emph{ACM Transactions on Graphics},
  vol.~36, no.~4, pp. 97--1, 2017.

\bibitem{teed2020raft}
Z.~Teed and J.~Deng, ``Raft: Recurrent all-pairs field transforms for optical
  flow,'' in \emph{European Conference on Computer Vision (ECCV)}.\hskip 1em
  plus 0.5em minus 0.4em\relax Springer, 2020, pp. 402--419.

\bibitem{sun2018multi}
S.-H. Sun, M.~Huh, Y.-H. Liao, N.~Zhang, and J.~J. Lim, ``Multi-view to novel
  view: Synthesizing novel views with self-learned confidence,'' in
  \emph{European Conference on Computer Vision (ECCV)}, 2018, pp. 155--171.

\bibitem{collins1996space}
R.~T. Collins, ``A space-sweep approach to true multi-image matching,'' in
  \emph{IEEE Conference on Computer Vision and Pattern Recognition
  (CVPR)}.\hskip 1em plus 0.5em minus 0.4em\relax IEEE, 1996, pp. 358--363.

\bibitem{simonyan2015very}
K.~Simonyan and A.~Zisserman, ``Very deep convolutional networks for
  large-scale image recognition,'' in \emph{International Conference on
  Learning Representations (ICLR)}, 2015.

\bibitem{wang2004image}
Z.~Wang, A.~C. Bovik, H.~R. Sheikh, and E.~P. Simoncelli, ``Image quality
  assessment: from error visibility to structural similarity,'' \emph{IEEE
  Transactions on Image Processing}, vol.~13, no.~4, pp. 600--612, 2004.

\bibitem{guo2021deep}
M.~Guo, J.~Hou, J.~Jin, J.~Chen, and L.-P. Chau, ``Deep spatial-angular
  regularization for light field imaging, denoising, and super-resolution,''
  \emph{IEEE Transactions on Pattern Analysis and Machine Intelligence}, 2021.

\bibitem{shi2019framework}
J.~Shi, X.~Jiang, and C.~Guillemot, ``A framework for learning depth from a
  flexible subset of dense and sparse light field views,'' \emph{IEEE
  Transactions on Image Processing}, vol.~28, no.~12, pp. 5867--5880, 2019.

\bibitem{wu2021revisiting}
G.~Wu, Y.~Liu, L.~Fang, and T.~Chai, ``Revisiting light field rendering with
  deep anti-aliasing neural network,'' \emph{IEEE Transactions on Pattern
  Analysis and Machine Intelligence}, 2021.

\bibitem{bao2019depth}
W.~Bao, W.-S. Lai, C.~Ma, X.~Zhang, Z.~Gao, and M.-H. Yang, ``Depth-aware video
  frame interpolation,'' in \emph{IEEE Conference on Computer Vision and
  Pattern Recognition (CVPR)}, 2019, pp. 3703--3712.

\bibitem{he2016deep}
K.~He, X.~Zhang, S.~Ren, and J.~Sun, ``Deep residual learning for image
  recognition,'' in \emph{IEEE Conference on Computer Vision and Pattern
  Recognition (CVPR)}, 2016, pp. 770--778.

\bibitem{riegler2021stable}
G.~Riegler and V.~Koltun, ``Stable view synthesis,'' in \emph{IEEE Conference
  on Computer Vision and Pattern Recognition (CVPR)}, 2021, pp.
  12\,216--12\,225.

\bibitem{kingma2015adam}
D.~P. Kingma and J.~L. Ba, ``Adam: A method for stochastic gradient descent,''
  in \emph{International Conference on Learning Representations (ICLR)}, 2015,
  pp. 1--15.

\bibitem{Vamsi2017}
V.~K. Adhikarla, M.~Vinkler, D.~Sumin, R.~Mantiuk, K.~Myszkowski, H.-P. Seidel,
  and P.~Didyk, ``Towards a quality metric for dense light fields,'' in
  \emph{IEEE Conference on Computer Vision and Pattern Recognition (CVPR)},
  2017, pp. 58--67.

\bibitem{aanaes2016large}
H.~Aan{\ae}s, R.~R. Jensen, G.~Vogiatzis, E.~Tola, and A.~B. Dahl,
  ``Large-scale data for multiple-view stereopsis,'' \emph{International
  Journal of Computer Vision}, vol. 120, no.~2, pp. 153--168, 2016.

\bibitem{wang2015occlusion}
T.-C. Wang, A.~A. Efros, and R.~Ramamoorthi, ``Occlusion-aware depth estimation
  using light-field cameras,'' in \emph{IEEE International Conference on
  Computer Vision (ICCV)}, 2015, pp. 3487--3495.

\bibitem{knapitsch2017tanks}
A.~Knapitsch, J.~Park, Q.-Y. Zhou, and V.~Koltun, ``Tanks and temples:
  Benchmarking large-scale scene reconstruction,'' \emph{ACM Transactions on
  Graphics}, vol.~36, no.~4, pp. 1--13, 2017.

\end{thebibliography}
%



%





\end{document}